\pdfoutput=1
\documentclass[journal]{IEEEtran}

\usepackage{cite}
\usepackage{url}
\usepackage{amsmath,amssymb,amsfonts}
\usepackage{graphicx}
\usepackage{textcomp}
\usepackage{tikz}
\usetikzlibrary{shapes.geometric, arrows.meta, positioning, fit}
\usepackage{pgfplots}
\pgfplotsset{compat=1.17}
\usepackage{booktabs}
\usepackage{multirow}
\usepackage{float}
\usepackage{placeins}

\newcommand{\history}[1]{}
\newcommand{\doi}[1]{}
\newcommand{\tfootnote}[1]{}
\newcommand{\corresp}[1]{}

\newenvironment{keywords}{\begin{IEEEkeywords}}{\end{IEEEkeywords}}
\newcommand{\EOD}{}

\AtBeginDocument{%
  \setcounter{dbltopnumber}{1}%
}

\begin{document}

\title{\textit{\small This work has been submitted to Taylor and Francis' Applied Artificial Intelligence for possible publication.}\\[0.5em]Multi-Dimensional Behavioral Evaluation of Agentic Stock Prediction Systems Using Large Language Model Judges with Closed-Loop Reinforcement Learning Feedback}
\author{\IEEEauthorblockN{Mohammad Al Ridhawi, Mahtab Haj Ali, and Hussein Al Osman}\\
\IEEEauthorblockA{School of Electrical Engineering and Computer Science,\\University of Ottawa, Ottawa, Canada\\e-mail: malri039@uottawa.ca}}

\markboth
{}
{}

\maketitle

\begin{abstract}
Agentic artificial intelligence systems produce outputs through sequences of interdependent autonomous decisions, yet standard evaluation assesses outputs alone and cannot diagnose the underlying process. We develop a behavioral evaluation methodology that complements output-level testing by scoring the intermediate decision process itself. Behavioral traces logged at each autonomous decision point are grouped into five-day episodes and scored along six domain-specific dimensions (regime detection, routing, adaptation, risk calibration, strategy coherence, error recovery) by an ensemble of three large language model (LLM) judges. A perturbation procedure that corrupts one dimension while leaving the other five intact confirms dimension specificity; cross-model agreement reaches Krippendorff's $\alpha = 0.85$. The composite behavioral score correlates at Spearman $\rho = 0.72$ with realized 20-day Sharpe ratio. Closing the loop, the framework converts deficient per-dimension scores into a credit-assigned penalty added to the Soft Actor-Critic reward. Three fine-tuning cycles, confined to validation data, reduce one-day MAPE from 0.61\% to 0.54\% (11.5\% relative; $p<0.001$, $d=0.31$) on the held-out 2017 to 2025 test period, significant under Diebold-Mariano and localized by Giacomini-White to the high-volatility regime. The methodology is application-agnostic and applies to any agentic system whose intermediate decisions can be logged.
\end{abstract}


%========================================================================
\section{Introduction}\label{sec:intro}
%========================================================================

Modern artificial intelligence systems are increasingly agentic in the operational sense that each output is produced through a sequence of interdependent autonomous decisions rather than through a single forward pass over a static model. A regime detector classifies the current state of the world, a routing layer selects between specialized predictive pathways, a controller adjusts hyperparameters in response to recent performance, and the final output emerges only after several such decisions have been chained \cite{wang2024survey,xi2023rise}. The autonomy and modularity that make these systems effective also make them difficult to evaluate. Standard evaluation practice scores the final output against a ground-truth label or a benchmark, but cannot localize which intermediate decision was responsible when the output is wrong, nor confirm that a correct output was produced through sound intermediate decisions rather than fortunate coincidence.

The distinction between output quality and process quality matters because the two can diverge. An agent may complete a task through a sequence of lucky coincidences in which several poor decisions cancel each other out, or fail a task despite making well-justified decisions that are then undermined by environmental stochasticity. As agentic systems are deployed in safety-relevant and high-cost domains (autonomous driving, medical decision support, financial forecasting, industrial control), the cost of treating output success as a proxy for decision quality grows. \cite{wang2024survey} survey the autonomous-agent literature and identify process-quality evaluation as one of the open problems for the field, noting that existing benchmarks measure whether agents achieve their goals but not whether their reasoning was sound. \cite{xi2023rise} similarly survey large-language-model based agents and observe that behavioral evaluation, in domains where the link between immediate outcome and decision quality is noisy, remains underdeveloped relative to output-level assessment.

The LLM-as-a-Judge paradigm offers a natural entry point for closing the gap. Large language models produce evaluations that align closely with human raters on natural-language tasks \cite{zheng2023judging,liu2023geval} and are reliable enough to drive downstream optimization in language-model alignment \cite{dubois2024alpacafarm,ouyang2022training}. Existing applications, however, evaluate static outputs such as single responses, summaries, and dialogue turns. Adapting the paradigm to agentic systems requires assessing temporal sequences of interdependent decisions made under stochastic conditions, where the same decision may be appropriate under one set of conditions and inappropriate under another, and where decision quality cannot always be inferred from the immediate outcome because of data-generating-process noise. To our knowledge, no prior work formalizes LLM-based behavioral evaluation as a methodology in which the dimensions, the validation procedure, and the integration with downstream learning are designed jointly against the requirements of a specific applied-AI task rather than borrowed wholesale from natural-language evaluation.

This paper develops such a methodology and demonstrates it as a case study on a representative agentic forecasting system: an adaptive regime-aware stock prediction system that combines an autoencoder regime detector, dual node-transformer pathways specialized for stable and volatile market conditions, and a Soft Actor-Critic (SAC) controller that tunes the regime threshold and the pathway blending weight at each trading day \cite{alridhawi2026regime,haarnoja2018soft}. The system is a representative instance of the agentic class because each daily forecast is produced through a four-stage decision chain (detect, route, predict, adjust), each stage is autonomous, and the quality of the forecast depends on the appropriateness of every stage's decision in context. The forecasting application is the demonstration domain rather than the contribution: the architecture is taken from prior work \cite{alridhawi2026nodetransformer,alridhawi2026regime} and is treated here as the agentic system being evaluated.

The methodology rests on four components: a behavioral-trace formalization that records the inputs, outputs, and adjustments at every autonomous decision point and segments them into fixed-length episodes (Equation~\ref{eq:trace}); six domain-specific evaluation dimensions covering three architectural decision points (regime detection, routing, adaptation) and three emergent behavioral properties (risk calibration, strategy coherence, error recovery) scored under an anchored Likert rubric; a perturbation procedure that corrupts exactly one dimension while leaving the other five mechanically intact, supplying direct validation of dimension specificity rather than indirect inference; and a credit-assignment mechanism that translates per-dimension diagnostics into targeted modifications of the reinforcement-learning reward, so that the existing controller corrects identified weaknesses without architectural changes. Empirical validation comprises perturbation-based dimension specificity, predictive-validity correlation against realized 20-day Sharpe ratio, and a closed-loop campaign whose effect on the underlying forecasts is checked against the baseline through paired tests, Diebold-Mariano \cite{diebold1995comparing,harvey1997testing}, Giacomini-White \cite{giacomini2006tests}, and Hansen model confidence set \cite{hansen2011model} procedures, plus canonical-benchmark comparisons.

The case study uses daily forecasts on 20 S\&P~500 equities over 1982 to 2025, with the closed-loop intervention evaluated on the held-out 2017 to 2025 test period (2{,}267 trading days). The dataset, feature engineering pipeline, and core evaluation metrics follow the protocol established in our prior work \cite{alridhawi2026regime}.

%========================================================================
\section{Materials and Methods}\label{sec:methods}
%========================================================================

The methodology developed here complements, rather than replaces, classical output-level evaluation. Procedures such as Diebold-Mariano \cite{diebold1995comparing,harvey1997testing}, Giacomini-White conditional predictive ability testing \cite{giacomini2006tests}, model confidence sets \cite{hansen2011model}, and forecast combination \cite{timmermann2006forecast} continue to supply the output-level evidence that the present framework supplements with process-level evidence. The principle that LLM judgments can serve as training signals for downstream learning is taken from reinforcement learning from human feedback \cite{ouyang2022training,dubois2024alpacafarm}; what is novel here is the adaptation of that principle to temporal sequences of interdependent decisions under stochastic conditions, the perturbation engineering that mechanically isolates each evaluation dimension, and the credit-assignment mechanism that routes per-dimension diagnostics to specific components of the controller's action space. The agent-evaluation literature has identified process quality as an open challenge \cite{yao2023react,shinn2023reflexion,wang2024survey,xi2023rise}, but a dedicated framework combining mechanical dimension-specificity validation, predictive-validity calibration, and a closed-loop bridge to a reinforcement-learning controller has not, to our knowledge, been established for any applied-AI domain. In financial forecasting specifically, evaluation has remained at the level of aggregate accuracy metrics and statistical predictive-accuracy tests \cite{lopezdeprado2018advances,aksehir2024analyzing,bao2025data}, leaving the quality of intermediate decisions unexamined.

\subsection{Case-Study System and Behavioral Context}\label{sec:system}

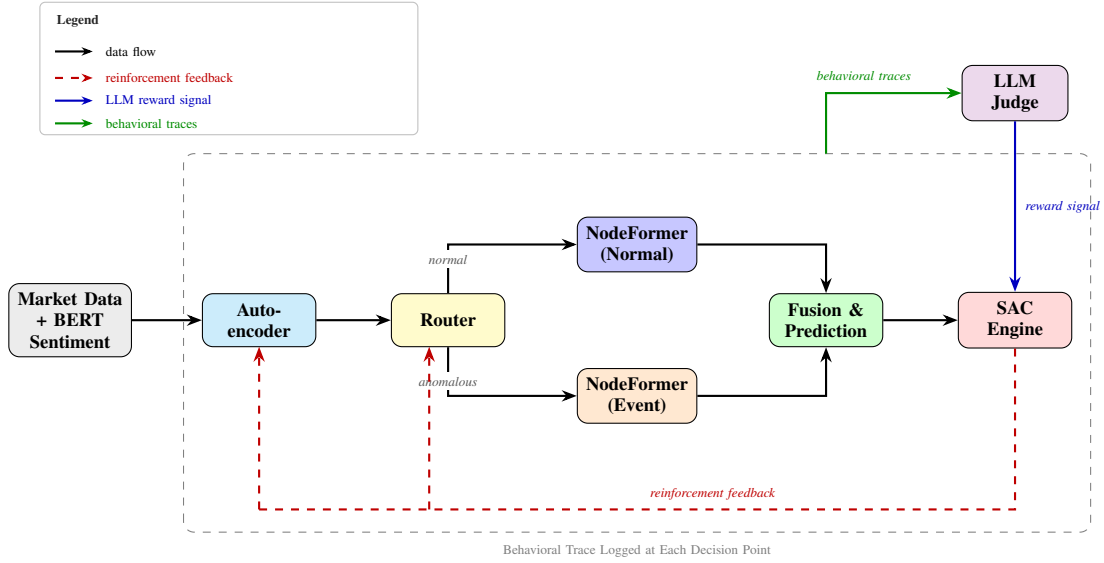
\begin{figure*}[tp]
\centering
\begin{tikzpicture}[
    node distance=0.5cm and 0.8cm,
    block/.style={rectangle, draw, rounded corners, fill=blue!8, minimum height=0.7cm, minimum width=1.5cm, align=center, font=\scriptsize\bfseries},
    arrow/.style={-{Stealth[length=2mm]}, thick},
    lbl/.style={font=\tiny\itshape, text=gray!70!black, fill=white, inner sep=1.5pt},
]
\node[block, fill=gray!15] (input) at (0,0) {Market Data\\+ BERT\\Sentiment};
\node[block, fill=cyan!18] (ae) at (2.5,0) {Auto-\\encoder};
\node[block, fill=yellow!25] (router) at (5,0) {Router};
\node[block, fill=blue!22] (nfn) at (7.5,1.0) {NodeFormer\\(Normal)};
\node[block, fill=orange!18] (nfa) at (7.5,-1.0) {NodeFormer\\(Event)};
\node[block, fill=green!20] (fusion) at (10,0) {Fusion \&\\Prediction};
\node[block, fill=red!15] (sac) at (12.5,0) {SAC\\Engine};
\draw[arrow] (input) -- (ae);
\draw[arrow] (ae) -- (router);
\draw[arrow] (router) |- (nfn) node[lbl, pos=0.35] {normal};
\draw[arrow] (router) |- (nfa) node[lbl, pos=0.35] {anomalous};
\draw[arrow] (nfn) -| (fusion);
\draw[arrow] (nfa) -| (fusion);
\draw[arrow] (fusion) -- (sac);
\draw[thick, dashed, red!75!black] (sac.south) -- (12.5,-2.5) -- (2.5,-2.5);
\draw[-{Stealth[length=2mm]}, thick, dashed, red!75!black] (4.75,-2.5) -- (4.75,-0.35);
\draw[-{Stealth[length=2mm]}, thick, dashed, red!75!black] (2.5,-2.5) -- (2.5,-0.35);
\node[font=\tiny\itshape, text=red!75!black] at (8.5,-2.3) {reinforcement feedback};
\node[block, fill=violet!15, minimum width=1.4cm] (llmfb) at (12.5,3.0) {LLM\\Judge};
\draw[arrow, thick, blue!75!black] (llmfb) -- (sac) node[lbl, midway, right, text=blue!75!black, xshift=2pt] {reward signal};
\draw[dashed, gray, rounded corners] (1.5,-2.8) rectangle (13.5,2.2);
\node[font=\tiny, gray, anchor=north] at (7.5,-2.85) {Behavioral Trace Logged at Each Decision Point};
\draw[arrow, thick, green!55!black] (10.0,2.2) |- (llmfb.west);
\node[font=\tiny\itshape, text=green!55!black, anchor=south] at (10.5,3.05) {behavioral traces};
\draw[draw=gray!50, rounded corners=2pt, fill=white, thin] (-0.4,2.45) rectangle (4.6,4.20);
\node[font=\tiny\bfseries, anchor=north west, text=gray!30!black] at (-0.3,4.15) {Legend};
\draw[-{Stealth[length=1.5mm]}, thick] (-0.2,3.55) -- (0.3,3.55);
\node[font=\tiny, anchor=west] at (0.35,3.55) {data flow};
\draw[-{Stealth[length=1.5mm]}, thick, dashed, red!75!black] (-0.2,3.20) -- (0.3,3.20);
\node[font=\tiny, anchor=west, text=red!75!black] at (0.35,3.20) {reinforcement feedback};
\draw[-{Stealth[length=1.5mm]}, thick, blue!75!black] (-0.2,2.90) -- (0.3,2.90);
\node[font=\tiny, anchor=west, text=blue!75!black] at (0.35,2.90) {LLM reward signal};
\draw[-{Stealth[length=1.5mm]}, thick, green!55!black] (-0.2,2.60) -- (0.3,2.60);
\node[font=\tiny, anchor=west, text=green!55!black] at (0.35,2.60) {behavioral traces};
\end{tikzpicture}
\caption{Architecture of the agentic prediction system with the LLM judge feedback loop. Behavioral traces logged at each decision point (dashed gray box) feed the LLM judge, which produces diagnostic reward signals for the SAC engine; the SAC engine also receives reinforcement feedback from realized prediction outcomes.}
\label{fig:architecture}
\end{figure*}

The framework is demonstrated on the adaptive regime-aware prediction system introduced in \cite{alridhawi2026regime}, which processes daily data for 20 S\&P~500 equities from January 1982 to March 2025. The complete architectural description and baseline evaluation are given in \cite{alridhawi2026regime}; only the elements the behavioral evaluation reads off are summarized here. At each trading day~$t$, the autoencoder computes reconstruction error $e_t = \|x_t - \hat{x}_t\|_2$ and classifies the market state by comparing $e_t$ to a learned threshold $\tau$; the router directs data to a normal- or event-conditioned pathway accordingly; the system emits a blended forecast $\hat{y}_t = \alpha_t\, y_{\text{normal}} + (1-\alpha_t)\, y_{\text{event}}$; and the Soft Actor-Critic (SAC) controller emits adjustments $(\Delta\tau_t, \Delta\alpha_t)$ in response to recent prediction performance. The forecast is therefore the output of a four-stage decision chain. Figure~\ref{fig:architecture} shows the complete pipeline; dashed red arrows mark the SAC reinforcement feedback loop from \cite{alridhawi2026regime}, and the blue arrow marks the new LLM diagnostic reward signal introduced here.

\subsection{Behavioral Trace Formalization}\label{sec:trace}

At each trading day $t$, the system records a behavioral trace $\mathcal{B}_t$ that captures the complete decision context and outcomes of every autonomous action. The trace is defined as a structured tuple:

\begin{equation}
\mathcal{B}_t = \langle\, \mathbf{m}_t,\; \mathbf{a}_t,\; \mathbf{r}_t,\; \mathbf{u}_t,\; \hat{y}_t,\; \mathbf{p}_t \,\rangle
\label{eq:trace}
\end{equation}

\noindent where the market context vector $\mathbf{m}_t = (p_t, v_t, \text{VIX}_t, \bar{S}_t)$ records the closing price, trading volume, CBOE Volatility Index value, and aggregated BERT sentiment score (an exponentially weighted average of BERT sentiment signals derived from financial news headlines, as in our prior work \cite{alridhawi2026nodetransformer}); the autoencoder decision vector $\mathbf{a}_t = (e_t, \tau_t, \ell_t)$ records the reconstruction error, the current threshold, and the binary regime label ($\ell_t \in \{0,1\}$, where 1 indicates anomalous); the routing vector $\mathbf{r}_t = (\alpha_t, k_t)$ records the blending weight and the dominant-pathway index ($k_t = 0$ normal, $k_t = 1$ event); the SAC action vector $\mathbf{u}_t = (\Delta\tau_t, \Delta\alpha_t) \in [-0.1, 0.1]^2$ records the threshold and blending-weight adjustments; $\hat{y}_t$ is the one-day-ahead forecast; and the rolling performance vector $\mathbf{p}_t = (\text{MAPE}_{t,w}, \text{DA}_{t,w})$ records MAPE and directional accuracy over a trailing window of $w = 20$ trading days. Including both $e_t$ and $\tau_t$ alongside $\ell_t$ allows the judge to assess the margin of the classification decision rather than only its label; consistency between $\ell_t$ and $(k_t, \alpha_t)$ is a key indicator of strategy coherence, since classifying the market as anomalous while setting $\alpha_t$ to heavily favor the normal pathway constitutes a logical contradiction. Adjustments are assessed for timeliness (applied soon after a detectable condition change), proportionality (scaled to the magnitude of the change), and stability (no day-to-day reversals).

The trace is serialized as a structured JSON record for inclusion in the LLM prompt, with labeled fields, units, and value ranges; reconstruction errors appear alongside the threshold, blending weights alongside the regime label, and SAC adjustments alongside the resulting post-update parameters. A representative serialized trace is reproduced in Appendix~\ref{app:prompts}. Single-day traces are insufficient for several evaluation dimensions, motivating the episodic grouping described next.

\subsection{Evaluation Episodes}\label{sec:episodes}

Several evaluation dimensions require temporal context that extends beyond a single day: adaptation responsiveness depends on whether the system adjusts parameters promptly after a condition change, error recovery on the system's behavior over multiple days following a prediction error, and strategy coherence on whether sequential decisions are logically consistent. The framework therefore groups consecutive traces into evaluation episodes:

\begin{equation}
\mathcal{E}_t = \{\mathcal{B}_t, \mathcal{B}_{t+1}, \mathcal{B}_{t+2}, \mathcal{B}_{t+3}, \mathcal{B}_{t+4}\}
\label{eq:episode}
\end{equation}

The five-day window matches the recent-error history length $k = 5$ used in the SAC state representation of \cite{alridhawi2026regime}. Validation-period pilots over three, five, seven, and ten-day windows favored five days on both inter-episode score variance ($\sigma = 0.71$ against $0.94$ at three days) and cross-model agreement (Krippendorff's $\alpha = 0.81$ against $0.74$ at three days and $0.78$ at ten days), with longer windows degrading agreement because justifications increasingly averaged over heterogeneous regimes within the episode. Episodes are sampled without overlap to reduce statistical dependence between evaluated units (see Section~\ref{sec:setup}).

\subsection{Evaluation Dimensions}\label{sec:dimensions}

The framework defines six evaluation dimensions in two categories. The first three correspond to specific architectural decision points and assess whether individual components performed their designated function. The remaining three capture emergent behavioral properties that arise from the interaction of multiple components and cannot be attributed to any single subsystem.

\subsubsection{Architectural decision dimensions}

\emph{Regime Detection (RD)} assesses the accuracy of the autoencoder's market-state classification and the sensitivity of the threshold $\tau$. The judge examines the relationship between $e_t$, $\tau_t$, $\ell_t$, and concurrent market context: a sharp VIX rise with elevated $e_t$ but a ``normal'' label receives a low RD score. Because regime detection is the first decision in the chain, errors here propagate downstream through routing and prediction.

\emph{Routing Appropriateness (RT)} evaluates whether data was directed to the correct pathway given the detected regime, with particular attention to decisions near the threshold boundary where $e_t \approx \tau_t$. A high RT score requires the blending weight $\alpha_t$ to reflect the system's confidence: smooth intermediate weights ($\alpha_t \approx 0.5$) when the classification is uncertain, and stable weights across consecutive days absent a corresponding change in conditions.

\emph{Adaptation Responsiveness (AD)} measures the timeliness and proportionality of SAC parameter adjustments. A high AD score requires that $\tau$ and $\alpha$ move within one to two trading days of a detectable shift, with magnitudes proportional to the severity of the change; persistent stale parameters or aggressive overshoot both lower the score. A day on which the VIX doubles while $\Delta\tau = \Delta\alpha = 0$ indicates poor adaptation.

\subsubsection{Emergent behavioral dimensions}

\emph{Risk Calibration (RC)} assesses the appropriateness of the system's overall risk posture given the prevailing volatility environment. A well-calibrated system becomes more conservative during high-volatility periods (lower $\alpha_t$, more cautious parameter adjustments) and more confident during low-volatility periods, rather than running a static configuration that ignores volatility.

\emph{Strategy Coherence (SC)} evaluates the absence of contradictory actions across the decision chain. Classifying the market as anomalous while setting $\alpha_t$ to favor the normal pathway, detecting a regime transition while freezing SAC parameters, or increasing $\tau$ (making detection less sensitive) while simultaneously shifting $\alpha$ toward the event pathway are all coherence violations the judge is asked to flag.

\emph{Error Recovery (ER)} measures the speed and effectiveness of adaptation following prediction errors. A substantial error at time $t$ is detected when the daily MAPE exceeds the trailing 20-day average by more than one standard deviation:

\begin{equation}
\xi_t = \mathbb{1}\!\left[\,\text{MAPE}_t > \overline{\text{MAPE}}_{t,20} + \sigma_{\text{MAPE},t,20}\,\right]
\label{eq:error_detect}
\end{equation}

\noindent where $\overline{\text{MAPE}}_{t,20}$ and $\sigma_{\text{MAPE},t,20}$ are the mean and standard deviation of daily MAPE over the preceding 20 trading days. When $\xi_t = 1$, a high ER score requires SAC adjustments within two trading days and measurable subsequent improvement; persistent errors and overcorrections both lower the score. Table~\ref{tab:dimensions} summarizes the six dimensions.

\begin{table}[!t]
\centering
\caption{Six evaluation dimensions for the LLM-as-a-Judge assessment, organized by category.}
\label{tab:dimensions}
\scriptsize
\setlength{\tabcolsep}{3pt}
\begin{tabular}{@{}llp{4.2cm}@{}}
\toprule
\textbf{Category} & \textbf{Dimension} & \textbf{Assessment Criteria} \\
\midrule
\multirow{3}{*}{\parbox{1.5cm}{Architectural\\Decision}} & Regime Detection (RD) & Accuracy of autoencoder classification; threshold sensitivity; false positive and false negative rates \\
\addlinespace[4pt]
& Routing (RT) & Consistency with detected regime; blending weight appropriateness; stability near threshold boundary \\
\addlinespace[4pt]
& Adaptation (AD) & Timeliness of SAC adjustments; proportionality to condition changes; absence of overshooting \\
\midrule
\multirow{3}{*}{\parbox{1.5cm}{Emergent\\Behavioral}} & Risk Calibration (RC) & Appropriateness of risk posture given volatility; conservative behavior in high-VIX periods \\
\addlinespace[4pt]
& Strategy Coherence (SC) & Logical consistency across decision chain; absence of contradictory actions \\
\addlinespace[4pt]
& Error Recovery (ER) & Speed of adaptation after errors; effectiveness of corrective actions; absence of overcorrection \\
\bottomrule
\end{tabular}
\end{table}

Coarser four-dimension and finer eight-dimension variants were piloted on the validation set: the four-dimension variant conflated distinct failure modes that require different corrective actions, and the eight-dimension variant produced lower inter-judge agreement (mean $\alpha_K = 0.62$) without improving predictive validity. Across the 200 unperturbed evaluation episodes, the mean pairwise Spearman correlation among the six dimensions is $\bar{\rho} = 0.26$ and no pairwise correlation exceeds 0.45 (the highest, $\rho = 0.41$, between regime detection and routing, reflects the architectural dependence of routing on regime classification), indicating substantially independent behavioral variation.

\subsection{Scoring Rubric}\label{sec:rubric}

Each dimension is scored on a 1 to 5 Likert scale anchored by observable behavioral criteria. An integer scale was preferred over a finer 1 to 10 scale, which produced lower inter-judge agreement in preliminary experiments without adding diagnostic information. Table~\ref{tab:rubric} reports the anchor descriptions for scores 1 and 5; scores 2 through 4 interpolate (2 indicates predominantly flawed performance with occasional acceptable behavior, 3 acceptable performance with identifiable but non-disqualifying weaknesses, and 4 strong performance with only minor imperfections). The full five-level rubric for all six dimensions is reproduced verbatim in the LLM prompt and in Appendix~\ref{app:prompts}.

\begin{table}[!t]
\centering
\caption{Rubric anchor descriptions for each evaluation dimension. Scores 2 through 4 interpolate between the anchors.}
\label{tab:rubric}
\scriptsize
\setlength{\tabcolsep}{3pt}
\begin{tabular}{@{}lp{3.6cm}p{3.6cm}@{}}
\toprule
\textbf{Dim.} & \textbf{Score 1 (Fundamentally Flawed)} & \textbf{Score 5 (Exemplary)} \\
\midrule
RD & Systematic misclassification of regime; threshold unresponsive to volatility changes & All regime transitions identified within one trading day; threshold adjustments proportional to volatility \\
\addlinespace[4pt]
RT & Data routed to wrong pathway; blending weight contradicts regime classification & Routing consistently matches market state; smooth blending transitions near threshold \\
\addlinespace[4pt]
AD & Parameters frozen or wildly oscillating; no response to condition changes within episode & Prompt, proportional adjustments within one to two days; no overshooting or oscillation \\
\addlinespace[4pt]
RC & Risk posture inappropriate for volatility (e.g., aggressive in high-VIX period) & Conservative during high volatility, confident during low volatility; smooth transitions \\
\addlinespace[4pt]
SC & Multiple contradictory actions across decision chain within episode & All decisions form logically consistent sequence; no contradictions \\
\addlinespace[4pt]
ER & No corrective action within 2 days of error; errors persist or worsen & Corrective adjustments within 1 to 2 days; subsequent predictions show measurable improvement \\
\bottomrule
\end{tabular}
\end{table}

The anchors reference concrete behavioral indicators the judge can read off the trace rather than abstract quality descriptors, following the principle from \cite{liu2023geval} that structured criteria yield more reliable assessments than open-ended scoring instructions. Inter-judge agreement on dimension scores is quantified using Krippendorff's alpha, appropriate for ordinal data with more than two raters and corrected for chance agreement.

\subsection{LLM Judge Ensemble and Evaluation Pipeline}\label{sec:prompt}

The LLM judge receives a structured prompt with four components: a fixed system description of the prediction architecture and trace-field semantics; the five-trace episode $\{\mathcal{B}_t, \ldots, \mathcal{B}_{t+4}\}$ serialized as JSON with units, ranges, and derived quantities such as $e_t/\tau_t$; the full evaluation rubric for each of the six dimensions; and an output schema requiring an integer score per dimension, a natural-language justification grounded in specific trace fields, and an optional failure label drawn from a 12-label vocabulary that maps each below-threshold score onto a specific SAC action subspace. The full prompt text and the failure-label vocabulary are reproduced verbatim in Appendix~\ref{app:prompts}. Following \cite{liu2023geval}, the prompt instructs the judge to summarize episode-level market conditions, examine each decision point against the information available when it was made, identify inconsistencies or failures with references to specific trace fields, and only then assign scores; the structured chain of thought, shown by \cite{liu2023geval} to improve evaluation fidelity on NLP tasks, extends to multi-step behavioral assessment where the judge must track causal chains across decision points and days.

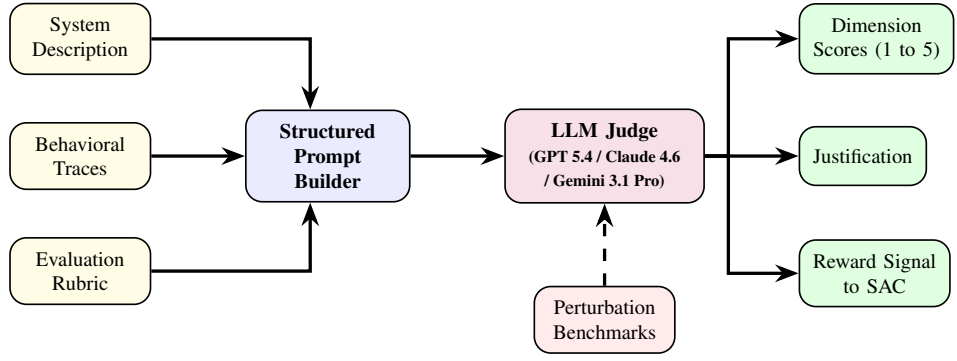
\begin{figure*}[tp]
\centering
\resizebox{0.70\textwidth}{!}{%
\begin{tikzpicture}[
    node distance=0.5cm and 0.6cm,
    block/.style={rectangle, draw, rounded corners, fill=blue!8, minimum height=0.6cm, minimum width=1.4cm, align=center, font=\tiny\bfseries},
    io/.style={rectangle, draw, rounded corners, fill=yellow!12, minimum height=0.5cm, minimum width=1.2cm, align=center, font=\tiny},
    output/.style={rectangle, draw, rounded corners, fill=green!12, minimum height=0.5cm, minimum width=1.0cm, align=center, font=\tiny},
    arrow/.style={-{Stealth[length=2mm]}, thick},
]
\node[io] (traces) {Behavioral\\Traces};
\node[io, below=0.4cm of traces] (rubric) {Evaluation\\Rubric};
\node[io, above=0.4cm of traces] (sysdesc) {System\\Description};
\node[block, right=0.8cm of traces] (prompt) {Structured\\Prompt\\Builder};
\node[block, right=0.8cm of prompt, fill=purple!12, minimum width=1.7cm] (llm) {LLM Judge\\{\fontsize{4pt}{5pt}\selectfont(GPT 5.4 / Claude 4.6}\\{\fontsize{4pt}{5pt}\selectfont / Gemini 3.1 Pro)}};
\node[output, right=0.8cm of llm, yshift=1.0cm] (scores) {Dimension\\Scores (1 to 5)};
\node[output, right=0.8cm of llm, yshift=0cm] (just) {Justification};
\node[output, right=0.8cm of llm, yshift=-1.0cm] (improve) {Reward Signal\\to SAC};
\draw[arrow] (sysdesc.east) -| ([xshift=-0.15cm]prompt.north);
\draw[arrow] (traces) -- (prompt);
\draw[arrow] (rubric.east) -| ([xshift=-0.15cm]prompt.south);
\draw[arrow] (prompt) -- (llm);
\draw[arrow] (llm.east) -- ++(0.2,0) |- (scores.west);
\draw[arrow] (llm.east) -- (just.west);
\draw[arrow] (llm.east) -- ++(0.2,0) |- (improve.west);
\node[io, below=0.7cm of llm, fill=red!8] (perturb) {Perturbation\\Benchmarks};
\draw[arrow, dashed] (perturb) -- (llm);
\end{tikzpicture}
}%
\caption{LLM-as-a-Judge evaluation pipeline. Behavioral traces, system description, and evaluation rubric are combined into a structured prompt. Outputs include per-dimension scores, natural-language justifications, and reward signals fed back into the SAC engine. Perturbation benchmarks validate the framework.}
\label{fig:pipeline}
\end{figure*}

Three judges from different model families evaluate every episode independently: GPT~5.4, Claude~4.6 Opus, and Gemini~3.1 Pro. Aggregating across families reduces the influence of any single model's idiosyncrasies and exposes inter-judge agreement as a per-dimension quality signal: higher agreement indicates the dimension is well-defined, while disagreement flags rubric ambiguity or genuine edge cases (Section~\ref{sec:agreement} reports the full agreement statistics and divergence examples). All judges operate at temperature zero, with no access to other judges' scores, and a maximum output of 2{,}000 tokens. Figure~\ref{fig:pipeline} illustrates the complete pipeline from trace collection to reward-signal generation.

The per-dimension consensus score $\bar{s}_d = \tfrac{1}{J}\sum_{j=1}^{J} s_{d,j}$ ($J = 3$) is the primary unit of evaluation. The closed-loop reward modification in Section~\ref{sec:closed_loop} operates on $\bar{s}_d$ directly with empirically calibrated weights $w_d$, preserving the diagnostic granularity that allows the SAC controller to localize which dimension requires correction. An unweighted composite $S_{\text{composite}} = \tfrac{1}{6}\sum_d \bar{s}_d$ is reported for cross-episode comparison but not used as a training signal, since collapsing the six dimensions into a scalar would discard targeted feedback.

\subsection{Closed-Loop Reward Integration}\label{sec:closed_loop}

\subsubsection{Reward modification}

The closed-loop mechanism augments the original SAC reward from \cite{alridhawi2026regime} (which combines prediction accuracy, directional correctness, and threshold stability) with a penalty derived from the LLM evaluation:

\begin{equation}
R'_t = R_t - \lambda \sum_{d=1}^{6} \max(0,\; \theta - s_d) \cdot w_d
\label{eq:reward}
\end{equation}

\noindent where $\lambda = 0.15$ scales LLM feedback against the primary reward, $\theta = 3$ is the minimum acceptable score (the midpoint of the 1 to 5 scale), and $w_d$ is the weight assigned to dimension $d$. Dimensions scoring at or above $\theta$ receive no penalty; the linear hinge ensures that severity scales proportionally with deficiency without the destabilizing magnitudes a quadratic or exponential penalty would produce. The feedback strength $\lambda$ was selected through a sweep on the validation set (2011 to 2016) over six values spanning an order of magnitude. For each candidate value, three full evaluation cycles were executed and the resulting metrics were recorded after the third cycle (Table~\ref{tab:lambda}).

\begin{table}[!t]
\centering
\caption{Sensitivity of closed-loop outcomes to the feedback strength $\lambda$, measured after three evaluation cycles on the validation set. $\bar{s}_{\text{def}}$ is the mean score across originally deficient dimensions.}
\label{tab:lambda}
\footnotesize
\begin{tabular}{@{}ccccl@{}}
\toprule
$\lambda$ & MAPE (\%) & DA (\%) & $\bar{s}_{\text{def}}$ & Stability \\
\midrule
0.05 & 0.60 & 72 & 2.8 & Stable \\
0.10 & 0.57 & 73 & 3.1 & Stable \\
0.15 & 0.54 & 74 & 3.6 & Stable \\
0.20 & 0.55 & 74 & 3.4 & Stable \\
0.25 & 0.58 & 72 & 3.0 & Marginal \\
0.30 & 0.63 & 71 & 2.5 & Unstable \\
\bottomrule
\end{tabular}
\end{table}

The sweep traces a three-regime pattern: under-correction at $\lambda \leq 0.05$ where the LLM penalty is dominated by the primary reward; an effective operating range at $\lambda \in [0.10, 0.20]$ where prediction and behavioral metrics improve jointly; and destabilization at $\lambda \geq 0.25$ where the reward signal causes oscillatory SAC behavior, with $\lambda = 0.30$ producing final metrics worse than the pre-intervention baseline. The selected $\lambda = 0.15$ is the point at which LLM feedback is strong enough to induce measurable behavioral change while remaining subordinate to the primary objective.

The dimension weights $w_d$ are set equal to the Spearman correlation between each dimension's score and the realized Sharpe ratio over the subsequent 20 trading days (Table~\ref{tab:correlation}), so the reward penalty emphasizes dimensions with the greatest demonstrated impact on risk-adjusted returns. The weights are computed once from the validation episodes and held fixed during the closed-loop cycles to prevent circular optimization in which the system simultaneously changes the behavior being evaluated and the weights used to evaluate it.

\subsubsection{Credit assignment}\label{sec:credit}

The SAC controller outputs a two-dimensional action $a_t = [\Delta\tau, \Delta\alpha]$, and different evaluation dimensions correspond to different action components. Routing the penalty incorrectly (for example, charging a routing deficiency to $\Delta\tau$ rather than $\Delta\alpha$) would create conflicting gradients and risk degrading non-targeted behaviors. The framework decomposes the penalty through a credit assignment vector $\mathbf{c}_d = [c_{d,\tau}, c_{d,\alpha}]$:

\begin{equation}
\Delta R_d = \max(0,\; \theta - \bar{s}_d) \cdot w_d \cdot \mathbf{c}_d
\label{eq:credit_penalty}
\end{equation}

\noindent with $\mathbf{c}_d = [1, 0]$ for dimensions targeting $\Delta\tau$ only (RD, RC), $[0, 1]$ for $\Delta\alpha$ only (RT), and $[0.5, 0.5]$ for both subspaces (AD, SC, ER); Table~\ref{tab:credit} gives the complete mapping.

\begin{table}[!t]
\centering
\caption{Credit assignment mapping from evaluation dimensions to SAC action subspaces.}
\label{tab:credit}
\scriptsize
\setlength{\tabcolsep}{3pt}
\begin{tabular}{@{}llp{3.6cm}@{}}
\toprule
\textbf{Dimension} & \textbf{Target Subspace} & \textbf{Rationale} \\
\midrule
Regime Detection & $\Delta\tau$ & Threshold directly controls regime classification boundary \\
Routing & $\Delta\alpha$ & Blending weight determines pathway contribution \\
Adaptation & $\Delta\tau$, $\Delta\alpha$ & Adaptation quality reflects both threshold and blending tuning \\
Risk Calibration & $\Delta\tau$ & Risk posture primarily determined by regime sensitivity \\
Strategy Coherence & $\Delta\tau$, $\Delta\alpha$ (uniform) & Coherence is a system-level property not localizable to one action \\
Error Recovery & $\Delta\tau$, $\Delta\alpha$ & Recovery requires coordinated adjustment of both parameters \\
\bottomrule
\end{tabular}
\end{table}

The total penalty from Equation~\ref{eq:reward} is thus applied selectively: each SAC action component receives penalty contributions only from dimensions with a non-zero element in its credit-assignment column. The ablation study in Section~\ref{sec:ablation} confirms that this targeted routing preserves the quality of non-deficient dimensions throughout the improvement cycles. The judge's structured JSON output also supplies, for each below-$\theta$ dimension, a categorical failure label (for example \texttt{delayed\_threshold}, \texttt{wrong\_routing}, \texttt{frozen\_parameters}, \texttt{inconsistent\_blend}); the label-to-subspace lookup in Appendix~\ref{app:prompts} provides a finer-grained routing signal that overrides the default dimensional mapping in atypical failure modes where, for instance, a routing deficiency originates from an incorrect threshold rather than the blending weight.

\subsubsection{Periodic evaluation protocol}\label{sec:protocol}

LLM evaluation operates in cycles rather than continuously, so the feedback acts as a corrective intervention rather than a competing reward stream. Each cycle begins with 40 trading days of normal operation (approximately two calendar months), generating traces without reward modification; the traces are grouped into eight non-overlapping five-day episodes, evaluated by all three judges, and per-dimension scores averaged across judges and episodes. If any mean dimension score falls below $\theta = 3$, the reward modification from Equation~\ref{eq:reward} is activated and the SAC controller is fine-tuned for 10 epochs on the cycle's replay buffer (Adam, learning rate $3 \times 10^{-4}$, soft target updates $\tau_{\text{soft}} = 0.005$, matching the original SAC training). The modified reward is deactivated after fine-tuning, and the system resumes normal operation. This periodic structure prevents the LLM feedback from dominating the reward signal during normal operation. Three cycles were conducted in the experiments of Section~\ref{sec:closed_loop_results}, at which point dimension scores converged above $\theta$ for all dimensions.

\subsection{Experimental Setup}\label{sec:setup}

The experiments use the same 20 S\&P~500 equities and temporal partitioning as our prior work \cite{alridhawi2026nodetransformer,alridhawi2026regime}: training over 1982 to 2010, validation over 2011 to 2016, and test over 2017 to 2025. The data discipline of \cite{alridhawi2026regime} is preserved: the test period is reserved for final reporting only and is never used for adaptation, hyperparameter tuning, or prompt selection. The closed-loop adaptation pipeline introduced here, including all SAC fine-tuning cycles, LLM-judge evaluation episodes, hyperparameter selection ($\lambda$ and $\theta$), prompt-sensitivity analyses, and ablations, is therefore confined entirely to the validation window. After the closed-loop process concludes, the SAC weights are frozen and the system is applied without further updates to the held-out test set for the final prediction-performance reporting in Section~\ref{sec:statistical_validation}.

For the LLM-judge analyses, 200 unperturbed evaluation episodes were sampled from the validation period, stratified by VIX-based regime: 70 episodes from low volatility (VIX $<$ 15), 90 from medium volatility (VIX 15 to 25), and 40 from high volatility (VIX $\geq$ 25). The distribution reflects the empirical frequency of each regime while over-sampling high volatility for statistical power, and episode boundaries were constrained to lie within a single regime classification. The validation window encompasses the European sovereign debt crisis (2011 to 2012), the 2014 to 2015 oil price crash, the 2015 Chinese market turbulence, and the 2016 Brexit period. All three judges evaluated the 200 episodes at temperature zero, independently. Total evaluation cost was approximately \$180; individual episode evaluations completed in 8 to 15 seconds per judge through provider batch endpoints. Table~\ref{tab:eval_config} summarizes the configuration.

\begin{table}[!t]
\centering
\caption{Evaluation framework configuration parameters.}
\label{tab:eval_config}
\scriptsize
\setlength{\tabcolsep}{3pt}
\begin{tabular}{@{}llp{3.3cm}@{}}
\toprule
\textbf{Parameter} & \textbf{Value} & \textbf{Rationale} \\
\midrule
Episode length & 5 trading days & Matches SAC history window ($k=5$) \\
Episode overlap & None & Statistical independence \\
Total episodes & 200 & Sufficient for regime stratification \\
Low-vol.\ episodes & 70 (VIX $<$ 15) & Proportional to regime frequency \\
Medium-vol.\ episodes & 90 (VIX 15 to 25) & Proportional to regime frequency \\
High-vol.\ episodes & 40 (VIX $\geq$ 25) & Over-sampled for statistical power \\
Evaluation period & 2011 to 2016 & Validation window (test held out) \\
LLM temperature & 0 & Deterministic, reproducible scores \\
Max output tokens & 2{,}000 & Accommodates CoT + 6 justifications \\
Score scale & 1 to 5 (integer) & Optimal LLM discrimination \\
Score threshold ($\theta$) & 3 & Midpoint of Likert scale \\
Feedback strength ($\lambda$) & 0.15 & Validated on 2011 to 2016 period \\
Cycle length & 40 trading days & Balance frequency / stability \\
SAC fine-tuning epochs & 10 per cycle & Sufficient for penalty integration \\
Perturbation episodes & 60 baseline $\times$ 6 types & 420 total perturbed + baseline \\
\bottomrule
\end{tabular}
\end{table}

\subsection{Perturbation Design}\label{sec:perturb_design}

To validate that the LLM judges can reliably detect specific behavioral deficiencies, the framework uses perturbation-based benchmarking. The validation set consists of 60 unperturbed baseline episodes (20 from each regime stratum) drawn from the 200 evaluation episodes. Six perturbations were designed, each engineered to corrupt exactly one dimension while leaving the other five mechanically unaffected. Letting $\mathcal{B}_t$ denote the original trace and $\mathcal{B}'_t$ the perturbed trace, the transformations are:

\begin{align}
\text{Regime inversion (RD):} \quad & \ell'_t = 1 - \ell_t, \quad e_t, \tau_t \text{ unchanged} \label{eq:perturb_rd} \\
\text{Wrong routing (RT):} \quad & \alpha'_t = 0.9 \cdot \ell_t + 0.1 \cdot (1 - \ell_t) \label{eq:perturb_rt} \\
\text{Frozen SAC (AD):} \quad & \mathbf{u}'_t = \mathbf{0} \;\;\forall\, t \in \mathcal{E} \label{eq:perturb_ad} \\
\text{No vol.\ scaling (RC):} \quad & f_{\text{VIX}}(\cdot) \to 1 \label{eq:perturb_rc} \\
\text{Contradictory (SC):} \quad & \exists\, i \neq j: \; a_i \perp a_j \text{ per day} \label{eq:perturb_sc} \\
\text{Disabled recovery (ER):} \quad & \mathbf{u}'_{t+1} = \mathbf{u}'_{t+2} = \mathbf{0} \;\;\text{if } \xi_t = 1 \label{eq:perturb_er}
\end{align}

Regime inversion (Eq.~\ref{eq:perturb_rd}) flips the classification while leaving $e_t$ and $\tau_t$ untouched, creating a label that contradicts the underlying evidence. Forced wrong-pathway routing (Eq.~\ref{eq:perturb_rt}) overrides the routing decision against the regime label with $\alpha = 0.1$ when normal and $\alpha = 0.9$ when anomalous, isolating RT from RD. Frozen SAC (Eq.~\ref{eq:perturb_ad}) zeros all adjustments for the episode, isolating AD. Removed VIX scaling (Eq.~\ref{eq:perturb_rc}) disables the volatility scaling in the risk-management code path, making the risk posture independent of current volatility. Randomized contradictory actions (Eq.~\ref{eq:perturb_sc}) replace one component per day with a value contradicting the others (anomalous label with $\alpha = 0.9$; or $\Delta\tau > 0$ paired with $\Delta\alpha$ shifted toward the event pathway), targeting coherence while preserving individual component outputs. Disabled recovery (Eq.~\ref{eq:perturb_er}) zeros SAC actions for two days after any $\xi_t = 1$ event, leaving non-error days untouched. Each perturbation was applied independently to the 60 baselines, yielding $6 \times 60 = 360$ perturbed episodes. Combined with the 60 originals, the validation set comprises 420 episodes; all three judges evaluated the complete set, producing $1{,}260$ individual results.

\subsection{Statistical Validation Procedures}\label{sec:stat_methods}

To assess whether closed-loop improvements transfer from validation to test data, the empirical analysis applies six complementary procedures on the held-out 2017 to 2025 test period: paired tests with Cohen's $d$ effect sizes \cite{cohen1988statistical}; bootstrap confidence intervals on the Sharpe-ratio improvement \cite{efron1993introduction}; the Diebold-Mariano test of equal predictive accuracy \cite{diebold1995comparing} with the \cite{harvey1997testing} small-sample correction and Newey-West HAC variance, replicated under squared-error, absolute-error, MAPE, and QLIKE losses to control the choice-of-loss confound; comparison against random-walk, AR(1), and ARMA(1,1)-GARCH(1,1) benchmarks; the Giacomini-White test of conditional predictive ability \cite{giacomini2006tests} with the autoencoder regime label as conditioning variable; and the Hansen model confidence set \cite{hansen2011model} across the closed-loop cycles and against a non-corrective SAC-only ablation. No weight updates, reward modifications, or hyperparameter adjustments occur on the test set under either configuration; Section~\ref{sec:statistical_validation} reports the single final comparison between pre- and post-intervention with SAC weights frozen after the third validation-period cycle.

%========================================================================
\section{Results}\label{sec:results}
%========================================================================

\subsection{Baseline System Performance}\label{sec:baseline}

Table~\ref{tab:baseline} reports one-day-ahead prediction performance for the agentic system and selected baselines prior to the LLM feedback loop, establishing the starting point from which the closed-loop improvement is measured. The full set of baseline comparisons, per-stock analyses, and multi-horizon results are reported in \cite{alridhawi2026regime}.

\begin{table}[!t]
\centering
\caption{One-day-ahead prediction performance before LLM feedback loop. DA: Directional Accuracy. CTR: Confidence Tracking Rate.}
\label{tab:baseline}
\footnotesize
\setlength{\tabcolsep}{3pt}
\begin{tabular}{@{}lccccc@{}}
\toprule
\textbf{Model} & \textbf{MAPE} & \textbf{RMSE} & \textbf{DA} & \textbf{Theil's U} & \textbf{CTR} \\
\midrule
ARIMA & 1.20\% & 1.35 & 55\% & 0.98 & 51\% \\
LSTM & 1.00\% & 1.20 & 58\% & 0.88 & 54\% \\
NodeFormer-BERT \cite{alridhawi2026nodetransformer} & 0.80\% & 0.95 & 65\% & 0.72 & 62\% \\
AE-NodeFormer (no SAC) & 0.68\% & 0.88 & 69\% & 0.68 & 64\% \\
AE-NodeFormer + SAC & \textbf{0.61\%} & \textbf{0.82} & \textbf{71\%} & \textbf{0.63} & \textbf{67\%} \\
\bottomrule
\end{tabular}
\end{table}

The AE-NodeFormer + SAC system from \cite{alridhawi2026regime}, applied without any closed-loop fine-tuning, achieves 0.61\% MAPE and 71\% directional accuracy on the validation-period episodes used here, a 24\% MAPE reduction over the single-pathway NodeFormer-BERT baseline from \cite{alridhawi2026nodetransformer}. These figures are within sampling noise of the test-period results in \cite{alridhawi2026regime} (0.59\% MAPE, 72\% DA across 2017 to 2025), confirming the validation window provides a representative operating regime. The remainder of this section compares the validation-period baseline against the closed-loop fine-tuned variant; held-out test-set prediction performance is reported separately in Section~\ref{sec:statistical_validation}. Five-day and 20-day MAPE follow the same pattern (1.02\% and 1.52\% respectively).

\subsection{Perturbation-Based Validation}\label{sec:perturb_results}

Table~\ref{tab:perturbation} reports the perturbation validation results. For each perturbation type, the table shows the targeted dimension, the mean score drop on that dimension (relative to the unperturbed baseline), the average score change on the remaining five dimensions (off-target shift), Krippendorff's alpha measuring cross-model agreement, and the statistical significance of the targeted drop.

\begin{table}[!t]
\centering
\caption{Perturbation validation results. $\Delta s$: mean targeted score drop (GPT~5.4). Off-target $\Delta$: average change on remaining five dimensions. $\alpha$: Krippendorff's cross-model agreement.}
\label{tab:perturbation}
\scriptsize
\setlength{\tabcolsep}{3pt}
\begin{tabular}{@{}llcccc@{}}
\toprule
\textbf{Perturbation} & \textbf{Target} & \textbf{$\Delta s$} & \textbf{Off-tgt.\ $\Delta$} & \textbf{$\alpha$} & \textbf{$p$-value} \\
\midrule
Regime inversion & RD & $-$2.4 & $-$0.3 & 0.85 & $<$0.001 \\
Wrong routing & RT & $-$2.1 & $-$0.4 & 0.81 & $<$0.001 \\
Frozen SAC & AD & $-$1.9 & $-$0.2 & 0.79 & $<$0.001 \\
No vol.\ scaling & RC & $-$1.7 & $-$0.3 & 0.74 & $<$0.001 \\
Contradictory action & SC & $-$2.2 & $-$0.5 & 0.82 & $<$0.001 \\
Disabled recovery & ER & $-$1.6 & $-$0.2 & 0.76 & $<$0.001 \\
\bottomrule
\end{tabular}
\end{table}

Every perturbation produces a statistically significant targeted score drop ($p < 0.001$, paired $t$-test). The mean drops range from $-1.6$ (disabled recovery) to $-2.4$ (regime inversion). Regime inversion produces the largest drop because the contradiction between the inverted label and the unchanged $(e_t, \tau_t, \text{VIX}_t)$ context is unambiguous across all three judges. Disabled recovery produces the smallest drop because in episodes where no significant errors occur within the five-day window the recovery behavior cannot be assessed in either direction, attenuating the mean effect.

\begin{figure*}[tp]
\centering
\begin{tikzpicture}
\begin{axis}[
    width=0.92\textwidth,
    height=6cm,
    ybar,
    bar width=9pt,
    xlabel={Perturbation Type},
    ylabel={Mean Score Change},
    ymin=-3.0, ymax=0,
    enlarge y limits=false,
    ytick={-3,-2,-1,0},
    xtick=data,
    symbolic x coords={Regime inv., Wrong rout., Frozen SAC, No vol.\ scl., Contrad.\ act., Disab.\ rec.},
    x tick label style={font=\tiny, rotate=25, anchor=east},
    legend style={at={(0.5,1.02)}, anchor=south, legend columns=2,
                  font=\scriptsize, draw=none, fill=none,
                  /tikz/every even column/.append style={column sep=1em}},
    font=\scriptsize,
    grid=major,
    grid style={gray!20},
    ymajorgrids=true,
    xmajorgrids=false,
    enlarge x limits=0.1,
    nodes near coords,
    every node near coord/.append style={font=\tiny},
]
\addplot[fill=red!40, draw=red!70] coordinates {
    (Regime inv., -2.4) (Wrong rout., -2.1) (Frozen SAC, -1.9)
    (No vol.\ scl., -1.7) (Contrad.\ act., -2.2) (Disab.\ rec., -1.6)
};
\addplot[fill=gray!30, draw=gray!60] coordinates {
    (Regime inv., -0.3) (Wrong rout., -0.4) (Frozen SAC, -0.2)
    (No vol.\ scl., -0.3) (Contrad.\ act., -0.5) (Disab.\ rec., -0.2)
};
\addlegendentry{Targeted $\Delta s$}
\addlegendentry{Off-target $\Delta$}
\end{axis}
\end{tikzpicture}
\caption{Targeted score drops (red) versus mean off-target shifts (gray) for each perturbation type. The large gap between targeted and off-target effects confirms dimension specificity: each perturbation degrades its intended dimension far more than other dimensions.}
\label{fig:perturbation}
\end{figure*}
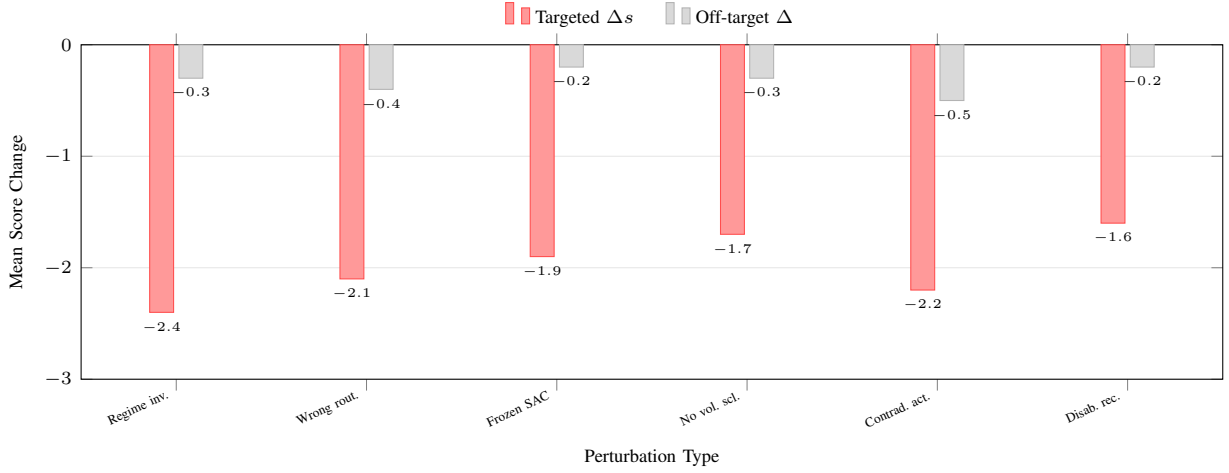

The mean off-target shift across all perturbations and dimensions is $0.32$, substantially smaller than the targeted drops. The largest off-target shift ($-0.5$) is for contradictory actions, since logically inconsistent decisions impair downstream quality on dimensions beyond coherence; the smallest ($-0.2$) are for frozen SAC and disabled recovery, indicating these perturbations are well-isolated. Figure~\ref{fig:perturbation} visualizes the contrast.

Non-zero off-target shifts are expected: a perturbation that degrades regime detection will have downstream effects on routing and adaptation, because subsequent decisions operate on incorrect regime information. Zero off-target sensitivity would in fact be suspect, implying that the judges assess each dimension in isolation rather than in the integrated manner behavioral evaluation requires. The observed pattern indicates that the judges recognize downstream effects while maintaining primary sensitivity to the targeted dimension.

\subsection{Cross-Model Agreement}\label{sec:agreement}

Cross-model agreement is quantified using Krippendorff's alpha, a reliability measure that generalizes across different numbers of raters and rating scales while correcting for chance agreement:

\begin{equation}
\alpha_K = 1 - \frac{D_o}{D_e} = 1 - \frac{(N-1)\sum_{c,k} o_{ck}\,\delta^2_{ck}}{\sum_{c,k} n_c\, n_k\, \delta^2_{ck}}
\label{eq:krippendorff}
\end{equation}

\noindent where $D_o$ and $D_e$ are observed and chance disagreement, $o_{ck}$ counts coincidences between categories $c$ and $k$, $n_c$ is the marginal frequency of category $c$, and $\delta^2_{ck}$ is the squared difference function appropriate for ordinal data. Thresholds of 0.667 and 0.800 are commonly accepted for tentative and reliable conclusions \cite{krippendorff2011computing}.

Across the six dimensions, $\alpha_K$ ranges from 0.74 (risk calibration) to 0.85 (regime detection). Regime detection, routing, and strategy coherence exceed 0.800. Regime detection achieves the highest agreement because the judges can directly compare $e_t$ against $\tau_t$ and verify the regime label against the market context; an obviously stable regime (low VIX, low volatility, no sentiment spikes) classified as anomalous is a clear error the three models identify consistently. Risk calibration achieves the lowest agreement because what constitutes an appropriate risk posture depends on implicit assumptions about risk tolerance, leading different LLM families to apply slightly different standards for conservative versus aggressive behavior. Even the lowest agreement (0.74) exceeds the threshold for meaningful inter-rater reliability, supporting use of all six dimensions in downstream reward modification.

Pairwise agreement shows GPT~5.4 and Claude~4.6 closest (Cohen's kappa \cite{cohen1960coefficient} 0.71 to 0.83 across dimensions); both show slightly lower agreement with Gemini~3.1 Pro (kappa 0.65 to 0.79). The divergence is not a uniform leniency or strictness bias: on identical episodes Gemini sometimes scored higher by treating borderline parameter lags as within acceptable response time, and sometimes lower by penalizing minor recovery overshoots the others treated as proportional (concrete examples in Section~\ref{sec:qualitative}). Averaging across the three judges yields more stable composite scores than any single judge.

\subsection{Per-Regime Score Analysis}\label{sec:regime_analysis}

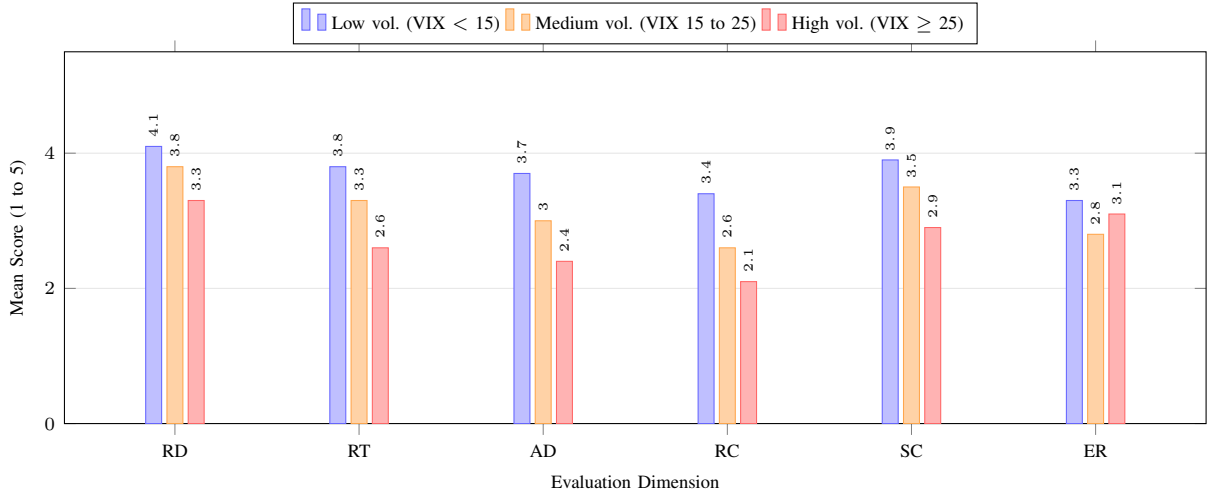
\begin{figure*}[tp]
\centering
\begin{tikzpicture}
\begin{axis}[
    width=0.92\textwidth,
    height=6.5cm,
    ybar,
    bar width=6pt,
    xlabel={Evaluation Dimension},
    ylabel={Mean Score (1 to 5)},
    ymin=0, ymax=5.5,
    xtick=data,
    symbolic x coords={RD, RT, AD, RC, SC, ER},
    legend style={at={(0.5,1.02)}, anchor=south, legend columns=3, font=\scriptsize},
    font=\scriptsize,
    grid=major,
    grid style={gray!20},
    ymajorgrids=true,
    xmajorgrids=false,
    enlarge x limits=0.12,
    nodes near coords,
    every node near coord/.append style={font=\tiny, rotate=90, anchor=west, xshift=0pt, yshift=0pt},
]
\addplot[fill=blue!25, draw=blue!60] coordinates {
    (RD, 4.1) (RT, 3.8) (AD, 3.7) (RC, 3.4) (SC, 3.9) (ER, 3.3)
};
\addplot[fill=orange!35, draw=orange!70] coordinates {
    (RD, 3.8) (RT, 3.3) (AD, 3.0) (RC, 2.6) (SC, 3.5) (ER, 2.8)
};
\addplot[fill=red!30, draw=red!60] coordinates {
    (RD, 3.3) (RT, 2.6) (AD, 2.4) (RC, 2.1) (SC, 2.9) (ER, 3.1)
};
\addlegendentry{Low vol.\ (VIX $<$ 15)}
\addlegendentry{Medium vol.\ (VIX 15 to 25)}
\addlegendentry{High vol.\ (VIX $\geq$ 25)}
\end{axis}
\end{tikzpicture}
\caption{Mean evaluation scores by volatility regime across the six dimensions (GPT~5.4, 200 unperturbed episodes). Risk calibration (RC) and adaptation responsiveness (AD) show the steepest declines from low to high volatility, while error recovery (ER) is non-monotonic.}
\label{fig:regime_scores}
\end{figure*}

To understand how behavioral quality varies across market conditions, Table~\ref{tab:regime_scores} reports the mean dimension scores from GPT~5.4 across the three regime strata on the 200 unperturbed evaluation episodes. GPT~5.4 is shown as a presentation choice (it has the highest pairwise agreement with the other two judges); all training and downstream analysis operate on the consensus score $\bar{s}_d$ averaged over the three judges, and the regime-dependent patterns produced by Claude~4.6 and Gemini~3.1 Pro are qualitatively similar. Figure~\ref{fig:regime_scores} visualizes the patterns.

\begin{table}[!t]
\centering
\caption{Mean dimension scores ($\pm$ standard error) by volatility regime on a 1 to 5 scale (GPT~5.4, unperturbed episodes; $n_{\text{low}} = 82$, $n_{\text{med}} = 74$, $n_{\text{high}} = 44$).}
\label{tab:regime_scores}
\scriptsize
\setlength{\tabcolsep}{2.5pt}
\begin{tabular}{@{}lcccccc@{}}
\toprule
\textbf{Regime} & \textbf{RD} & \textbf{RT} & \textbf{AD} & \textbf{RC} & \textbf{SC} & \textbf{ER} \\
\midrule
Low vol.\ (VIX $<$ 15) & 4.1\tiny{$\pm$0.07} & 3.8\tiny{$\pm$0.09} & 3.7\tiny{$\pm$0.11} & 3.4\tiny{$\pm$0.12} & 3.9\tiny{$\pm$0.08} & 3.3\tiny{$\pm$0.14} \\
Medium vol.\ (VIX 15 to 25) & 3.8\tiny{$\pm$0.09} & 3.3\tiny{$\pm$0.12} & 3.0\tiny{$\pm$0.14} & 2.6\tiny{$\pm$0.17} & 3.5\tiny{$\pm$0.11} & 2.8\tiny{$\pm$0.15} \\
High vol.\ (VIX $\geq$ 25) & 3.3\tiny{$\pm$0.12} & 2.6\tiny{$\pm$0.16} & 2.4\tiny{$\pm$0.18} & 2.1\tiny{$\pm$0.19} & 2.9\tiny{$\pm$0.14} & 3.1\tiny{$\pm$0.17} \\
\bottomrule
\end{tabular}
\end{table}

Behavioral quality generally degrades from low to high volatility, though not uniformly. Risk calibration and adaptation responsiveness decline most steeply (both by 1.3 points). Risk calibration drops from 3.4 (low) to 2.1 (high), so risk management is least adequate precisely when it matters most. Routing declines by 1.2 points, reflecting the difficulty of stable routing when $e_t$ fluctuates near $\tau$.

Error recovery is a notable exception: it scores higher in high-volatility episodes (3.1) than in medium-volatility episodes (2.8). The reversal occurs because high-volatility periods produce more frequent and pronounced prediction errors, giving the system clearer signals for corrective action and more opportunities to demonstrate recovery; during medium volatility, errors are less severe and the appropriate response is more ambiguous. Regime detection stays relatively high even in high volatility (3.3), indicating the autoencoder reliably identifies volatile conditions; the bottleneck under stress is the system's response to anomalous conditions, not their detection.

This pattern motivates the closed-loop mechanism: behavioral weaknesses are concentrated in specific dimensions under specific conditions, and a targeted intervention should produce disproportionate gains in those conditions. The prediction is confirmed in Section~\ref{sec:closed_loop_results}, where the largest improvements occur during high-volatility episodes.

\subsection{Predictive Validity}\label{sec:predictive_validity}

A behavioral evaluation framework is useful only if its scores predict future system performance. To assess predictive validity, Spearman rank correlations were computed between GPT~5.4's dimension scores on each unperturbed episode and realized trading metrics over the subsequent 20 trading days. The 20-day window captures medium-term consequences of behavioral quality without extending so far that the link to evaluated behavior is diluted by intervening events. Table~\ref{tab:correlation} reports the correlations.

\begin{table}[!t]
\centering
\caption{Spearman $\rho$ between GPT~5.4 dimension scores and realized 20-day trading metrics. MAPE$^{-1}$ and MDD$^{-1}$ are inverted so that higher values indicate better performance (all correlations significant at $p < 0.01$).}
\label{tab:correlation}
\footnotesize
\setlength{\tabcolsep}{4pt}
\begin{tabular}{@{}lcccc@{}}
\toprule
\textbf{Dimension} & \textbf{Sharpe} & \textbf{MAPE$^{-1}$} & \textbf{MDD$^{-1}$} & \textbf{Return} \\
\midrule
Regime Detection & 0.64 & \textbf{0.69} & 0.51 & 0.54 \\
Routing & 0.59 & 0.61 & 0.47 & 0.51 \\
Adaptation & 0.58 & 0.53 & 0.56 & 0.52 \\
Risk Calibration & 0.55 & 0.44 & \textbf{0.64} & 0.43 \\
Strategy Coherence & 0.62 & 0.59 & 0.49 & 0.57 \\
Error Recovery & 0.51 & 0.48 & 0.54 & 0.46 \\
\midrule
Composite & \textbf{0.72} & 0.68 & 0.61 & \textbf{0.59} \\
\bottomrule
\end{tabular}
\end{table}

The composite score correlates most strongly with the Sharpe ratio ($\rho = 0.72$), so overall behavioral quality predicts risk-adjusted returns over the subsequent month. Because the Sharpe ratio integrates both the magnitude and the consistency of returns, behavioral quality predicts not just average performance but performance stability.

Individual dimensions exhibit predictive profiles that align with their conceptual definitions. Regime detection is the strongest predictor of inverse MAPE ($\rho = 0.69$), consistent with accurate classification enabling appropriate model selection. Risk calibration is the strongest predictor of inverse maximum drawdown ($\rho = 0.64$), reflecting that appropriate risk posture during volatile periods limits downside exposure; that it predicts drawdown best despite showing the lowest inter-judge agreement suggests imperfect calibration assessments still capture genuine variation in tail-risk behavior. Strategy coherence shows the strongest individual correlation with returns ($\rho = 0.57$), so logical consistency across the decision chain translates into more reliable exploitation of predictive signals. Risk calibration's weak link to MAPE$^{-1}$ ($\rho = 0.44$) but strong link to drawdown indicates its contribution operates through risk management rather than forecasting. Error recovery's moderate correlations ($\rho$ from 0.46 to 0.54) reflect that recovery only manifests when errors occur, limiting its predictive content in error-free episodes. Even the weakest correlation is significant ($p < 0.01$).

These correlations serve a dual purpose: they validate that LLM scores reflect genuine behavioral properties influencing realized outcomes, and they supply the empirical basis for the dimension weights $w_d$ in Equation~\ref{eq:reward}, ensuring the closed-loop penalty emphasizes the dimensions with the greatest demonstrated impact on risk-adjusted performance.

\subsection{Qualitative Analysis of LLM Judgments}\label{sec:qualitative}

Four representative judgment patterns illustrate how the judges produce evidence-grounded assessments rather than generic quality ratings; the four span a stable episode, a transition episode, a perturbed episode, and an error-recovery episode, covering the high-, mid-, and low-agreement dimensions of Section~\ref{sec:agreement}. On a high-quality episode in stable conditions (VIX averaging 13.2), all three judges assigned 4 or 5 across all dimensions. GPT~5.4 noted that ``$e_t/\tau_t$ stays between 0.45 and 0.62, and $\alpha = 0.82$ to $0.88$ reflects high confidence in the normal-regime classification.'' Claude~4.6 highlighted that ``SAC adjustments are minimal ($|\Delta\tau| < 0.005$, $|\Delta\alpha| < 0.02$), appropriate given the absence of condition changes.''

On a challenging episode spanning a volatility spike (VIX 18 to 29 over five days), the judges assigned lower scores on risk calibration and adaptation. GPT~5.4 gave RC~=~2, noting that ``the system maintains $\alpha = 0.75$ on day 3 despite VIX exceeding 25, and the day-4 drop to $\alpha = 0.42$ lags the spike by a full trading day.'' Gemini~3.1 Pro gave RC~=~3 on the same episode, identifying the same lag but interpreting it as ``within acceptable response time given the noise in daily VIX.'' The divergence illustrates the subjective component of risk-calibration assessment that contributes to its lower Krippendorff's alpha.

On a perturbed episode with regime-label inversion, all three judges identified the perturbation with high confidence. Claude~4.6 gave RD~=~1: ``the regime label indicates `normal' on all five days despite $e \in [0.041, 0.053]$ exceeding $\tau = 0.031$, and VIX 27 to 32 that is inconsistent with normal conditions.'' The failure label \texttt{systematic\_misclassification} routes to $\Delta\tau$. Routing was only moderately affected (RT~=~3) because $\alpha = 0.55$ remained a reasonable intermediate value despite the inverted label, illustrating the off-target specificity pattern from the quantitative analysis.

A fourth example illustrates error recovery. During an episode in which $\xi_t = 1$ on days 1 and 2 (Eq.~\ref{eq:error_detect}), GPT~5.4 gave ER~=~4: ``the SAC controller reduces $\tau$ by 0.08 on day 3, $\alpha$ moves from 0.71 to 0.54 by day 4, and day-5 MAPE returns to within one standard deviation of the rolling mean. The corrective sequence is prompt and proportional.'' Claude~4.6 also gave ER~=~4; Gemini~3.1 Pro gave ER~=~3, noting ``the day-3 correction overshoots slightly, as $\tau$ drops 0.03 below its pre-error level.'' The disagreement reflects an interpretive difference about where proportional ends and overshoot begins when the overshoot is small. The episode also shows how recovery scores depend on errors being present: in the 82 low-volatility episodes with no significant errors, all three judges defaulted to ER~=~3, noting insufficient evidence.

These examples demonstrate that the judges ground their assessments in specific numerical values, articulate causal chains between upstream decisions and downstream consequences, agree on objective dimensions (RD, RT) where rubric criteria are unambiguous, and disagree on subjective dimensions (RC, ER) where multiple valid interpretations exist. The lower Krippendorff's alpha for ER thus reflects genuine interpretive ambiguity about proportionality rather than poor instrument design.

\subsection{Closed-Loop Improvement}\label{sec:closed_loop_results}

Three cycles of LLM evaluation and SAC reward modification were conducted with $\lambda = 0.15$. The data partitioning mirrors the discipline of \cite{alridhawi2026regime}: every adaptation step, hyperparameter choice, and fine-tuning update operates on validation-period data, and the test period is reserved for final reporting. The three cycles ran on a contiguous 120-day window in the latter portion of the validation set (three consecutive 40-day cycles ending in late 2016), with each cycle's SAC fine-tuning drawing only on its own replay buffer. After the third cycle, all SAC weights are frozen and the controller is applied without further updates to the held-out 2017 to 2025 test period in Section~\ref{sec:statistical_validation}; the SAC weights have therefore never been updated using transitions from the test set.

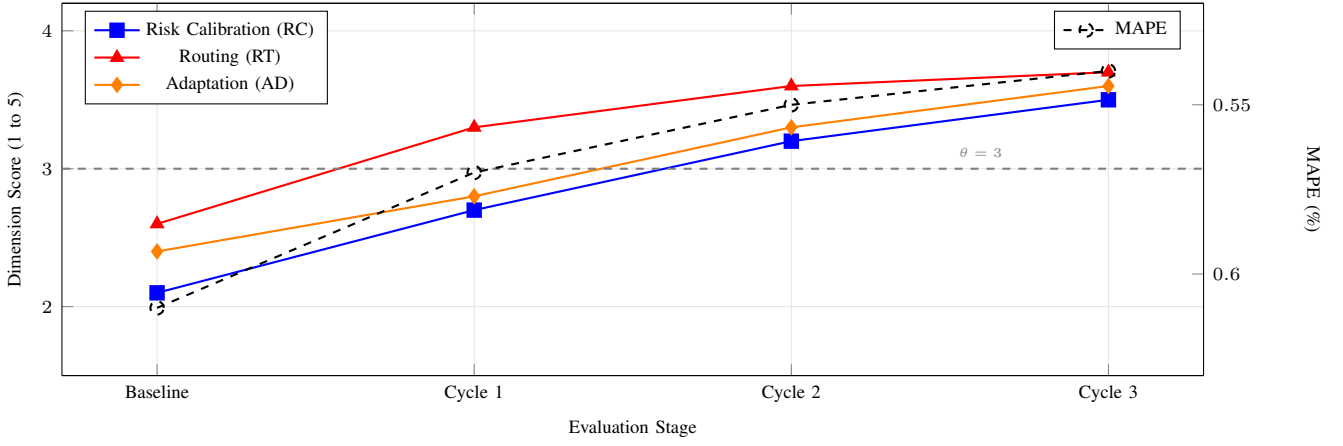
\begin{figure*}[tp]
\centering
\begin{tikzpicture}
\begin{axis}[
    width=0.92\textwidth,
    height=6.5cm,
    xlabel={Evaluation Stage},
    ylabel={Dimension Score (1 to 5)},
    xmin=-0.3, xmax=3.3,
    ymin=1.5, ymax=4.2,
    xtick={0,1,2,3},
    xticklabels={Baseline, Cycle 1, Cycle 2, Cycle 3},
    legend style={at={(0.02,0.98)}, anchor=north west, font=\scriptsize},
    font=\scriptsize,
    grid=major,
    grid style={gray!20},
    axis y line*=left,
]
\addplot[blue, thick, mark=square*, mark size=2.5] coordinates {
    (0, 2.1) (1, 2.7) (2, 3.2) (3, 3.5)
};
\addlegendentry{Risk Calibration (RC)}
\addplot[red, thick, mark=triangle*, mark size=2.5] coordinates {
    (0, 2.6) (1, 3.3) (2, 3.6) (3, 3.7)
};
\addlegendentry{Routing (RT)}
\addplot[orange, thick, mark=diamond*, mark size=2.5] coordinates {
    (0, 2.4) (1, 2.8) (2, 3.3) (3, 3.6)
};
\addlegendentry{Adaptation (AD)}
\draw[dashed, gray, thick] (axis cs:-0.3,3) -- (axis cs:3.3,3);
\node[font=\tiny, gray, anchor=south west] at (axis cs:2.5,3.02) {$\theta = 3$};
\end{axis}
\begin{axis}[
    width=0.92\textwidth,
    height=6.5cm,
    xmin=-0.3, xmax=3.3,
    ymin=0.52, ymax=0.63,
    xtick=\empty,
    axis y line*=right,
    ylabel={MAPE (\%)},
    ylabel style={rotate=180, at={(1.08,0.5)}},
    legend style={at={(0.98,0.98)}, anchor=north east, font=\scriptsize},
    font=\scriptsize,
    y dir=reverse,
]
\addplot[black, thick, dashed, mark=o, mark size=2.5] coordinates {
    (0, 0.61) (1, 0.57) (2, 0.55) (3, 0.54)
};
\addlegendentry{MAPE}
\end{axis}
\end{tikzpicture}
\caption{Progression of deficient dimension scores (left axis, solid lines) and MAPE (right axis, dashed line, inverted) across three closed-loop cycles, measured on the validation-period episodes within each cycle. The horizontal dashed gray line marks the intervention threshold ($\theta = 3$).}
\label{fig:closed_loop}
\end{figure*}

The initial evaluation identified three dimensions below $\theta = 3$: risk calibration during regime transitions (mean 2.1), routing near the threshold boundary (mean 2.6), and adaptation following consecutive errors (mean 2.4), with deficiencies most pronounced in high-volatility episodes. Table~\ref{tab:closed_loop} summarizes the per-cycle validation-period progression and Figure~\ref{fig:closed_loop} plots the trajectories.

\begin{table}[!t]
\centering
\caption{Per-cycle progression of deficient dimension scores ($\pm$ standard error) and prediction metrics during the closed-loop process. All entries are measured on validation-period episodes; the held-out test results are reported in Section~\ref{sec:statistical_validation}.}
\label{tab:closed_loop}
\scriptsize
\setlength{\tabcolsep}{3pt}
\begin{tabular}{@{}lcccc@{}}
\toprule
\textbf{Metric} & \textbf{Baseline} & \textbf{Cycle 1} & \textbf{Cycle 2} & \textbf{Cycle 3} \\
\midrule
Risk Calibration (RC) & 2.1\tiny{$\pm$0.18} & 2.7\tiny{$\pm$0.16} & 3.2\tiny{$\pm$0.14} & 3.5\tiny{$\pm$0.11} \\
Routing (RT, near $\tau$) & 2.6\tiny{$\pm$0.15} & 3.3\tiny{$\pm$0.12} & 3.6\tiny{$\pm$0.10} & 3.7\tiny{$\pm$0.09} \\
Adaptation (AD, post-error) & 2.4\tiny{$\pm$0.17} & 2.8\tiny{$\pm$0.15} & 3.3\tiny{$\pm$0.12} & 3.6\tiny{$\pm$0.10} \\
\midrule
Val.\ MAPE (\%) & 0.61\tiny{$\pm$0.03} & 0.57\tiny{$\pm$0.03} & 0.55\tiny{$\pm$0.02} & 0.54\tiny{$\pm$0.02} \\
Val.\ DA (\%) & 71\tiny{$\pm$2} & 73\tiny{$\pm$2} & 74\tiny{$\pm$2} & 74\tiny{$\pm$2} \\
\bottomrule
\end{tabular}
\end{table}

After three cycles, all three deficient dimensions exceed $\theta$. Risk calibration rose from 2.1 to 3.5: the SAC controller produces larger $\Delta\alpha$ adjustments toward the event pathway when $e_t$ rises, even before crossing $\tau$, reflecting a proactive rather than reactive response to emerging volatility. Routing rose from 2.6 to 3.7, the gain concentrated in the first cycle as the $\Delta\alpha$-targeted penalty took immediate effect; the system now sets intermediate weights ($\alpha \approx 0.4$ to $0.6$) when $e_t$ is within 15\% of $\tau$ rather than committing fully to one pathway. Adaptation rose from 2.4 to 3.6, with corrective adjustments applied earlier in error sequences rather than after error magnitudes accumulate.

These validation-period behavioral gains were tested on unseen data by applying the frozen post-cycle-3 SAC controller to the held-out 2017 to 2025 test period and comparing against the pre-intervention configuration on identical test inputs. Test-period one-day MAPE fell from 0.61\% to 0.54\% (11.5\% relative reduction), directional accuracy rose from 71\% to 74\% (3 percentage points), and the Sharpe ratio increased by 18\%. The gains are concentrated where the pre-intervention deficiencies were greatest: test-set MAPE in high-volatility episodes decreased by 17.3\% versus 7.8\% in low-volatility episodes. Longer horizons attenuate but remain significant: five-day MAPE decreased 10.8\% (1.02\% to 0.91\%) and twenty-day MAPE 8.6\% (1.52\% to 1.39\%), with directional accuracy gains of 2 and 1 points respectively.

The per-cycle progression in Table~\ref{tab:closed_loop} shows diminishing returns consistent with convergence, with the rate varying by dimension. Routing improved most rapidly (2.6 to 3.3 in cycle 1), risk calibration more gradually (requiring all three cycles to exceed $\theta$), and adaptation produced its largest gain in cycle 2 (2.8 to 3.3), likely because the cycle-1 routing improvements supplied a more useful state representation. By cycle 3 the magnitude of reward modifications had decreased substantially, indicating the controller had internalized the corrections. Dimensions already above $\theta$ at the start (RD 3.8, SC 3.4, ER 3.2) remained stable throughout (mean changes $<$ 0.2), confirming that the credit-assignment mechanism (Table~\ref{tab:credit}) routes the targeted penalty without degrading already-adequate behaviors.

\subsection{Statistical Validation}\label{sec:statistical_validation}

This subsection situates the closed-loop intervention within the standard predictive-accuracy testing framework, applying the six complementary procedures introduced in Section~\ref{sec:stat_methods} to the held-out 2017 to 2025 test period. Each procedure probes a different alternative explanation for the observed gain: paired tests and bootstrap intervals establish significance, Diebold-Mariano and benchmark comparisons rule out artefacts, conditional predictive ability and the Hansen confidence set probe regime and cycle robustness, and the four-loss sweep verifies the result is not specific to squared-error loss.

\paragraph{Paired significance and Sharpe bootstrap.} On the held-out 2017 to 2025 test period, paired $t$-tests compare daily prediction errors between the pre-intervention configuration (AE-NodeFormer + SAC from \cite{alridhawi2026regime}, no closed-loop fine-tuning) and the post-intervention configuration (same system after the three closed-loop cycles). Both run with all weights frozen on identical inputs, isolating the contribution of the closed-loop fine-tuning. The MAPE reduction (0.61\% $\to$ 0.54\%) is significant ($t = 4.82$, $p < 0.001$, Cohen's $d = 0.31$; \cite{cohen1988statistical}), as is the directional-accuracy gain (71\% $\to$ 74\%; $t = 3.15$, $p = 0.002$, $d = 0.20$). Effect sizes are moderate relative to the architectural innovations in \cite{alridhawi2026regime} ($d$ of 0.27 to 0.57), consistent with the closed-loop refining an already strong system. Five-day MAPE is significant ($t = 3.41$, $p < 0.001$, $d = 0.22$); twenty-day MAPE is significant at a lower confidence ($t = 2.28$, $p = 0.024$, $d = 0.15$), the declining effect plausibly due to compounding variance over multi-day windows. A bootstrap \cite{efron1993introduction} with 10{,}000 resamples yields a 95\% CI for the Sharpe-ratio improvement of [8.2\%, 27.4\%] around the 18\% point estimate, excluding zero.

\paragraph{Diebold-Mariano test of equal predictive accuracy.} The Diebold-Mariano (DM) test \cite{diebold1995comparing} with the \cite{harvey1997testing} small-sample correction is applied to per-day squared-error loss differentials $d_{i,t} = (e^{(\mathrm{pre})}_{i,t})^2 - (e^{(\mathrm{post})}_{i,t})^2$. The per-stock statistic is $\mathrm{DM}_i = \bar{d}_i / \sqrt{\hat{V}(\bar{d}_i)}$ with $\hat{V}$ a Newey-West HAC estimator (bandwidth $\lfloor 4(T/100)^{2/9} \rfloor = 5$). The fixed-effects panel pool over 20 stocks and 2{,}267 trading days yields $\mathrm{DM} = -7.83$ ($p < 0.001$), with the negative sign indicating strictly lower squared-error loss post-intervention. Per-stock statistics range from $-1.74$ (low-volatility consumer staples, smallest pre-intervention errors) to $-6.49$ (high-beta technology, largest pre-intervention errors). Eighteen of twenty per-stock tests remain significant after Bonferroni correction ($|z| = 2.81$); all twenty are significant at the uncorrected 5\% level. The improvement is therefore not driven by a small number of tickers.

\paragraph{Comparison against statistical benchmarks.} The agentic system is situated against three canonical univariate forecasters on the same test inputs: a no-change random walk; an AR(1) on log-returns with parameters re-estimated on a 1{,}000-day rolling window; and an ARMA(1,1)-GARCH(1,1) on the same window with the conditional mean used as the point forecast. The third rules out the alternative that agentic gains depend on conditional-variance modeling. Table~\ref{tab:benchmarks} summarizes the comparison.

\begin{table}[!t]
\centering
\caption{One-day-ahead test-period predictive accuracy against statistical benchmarks. Pooled Diebold-Mariano statistics compare each row to the post-intervention agentic system (final column) under squared-error loss; negative values indicate the agentic system has lower loss.}
\label{tab:benchmarks}
\scriptsize
\setlength{\tabcolsep}{3pt}
\begin{tabular}{@{}lcccc@{}}
\toprule
\textbf{Forecaster} & \textbf{MAPE} & \textbf{RMSE} & \textbf{DA} & \textbf{DM vs.\ post} \\
\midrule
Random Walk                 & 1.34\% & 1.51 & 50\% & $-23.74$ \\
AR(1) on log-returns        & 1.31\% & 1.48 & 51\% & $-22.41$ \\
ARMA(1,1)-GARCH(1,1)        & 1.29\% & 1.46 & 52\% & $-21.86$ \\
Pre-intervention agentic    & 0.61\% & 0.82 & 71\% & $-7.83$ \\
\textbf{Post-intervention}  & \textbf{0.54\%} & \textbf{0.74} & \textbf{74\%} & (reference) \\
\bottomrule
\end{tabular}
\end{table}

The post-intervention agentic system achieves a one-day MAPE roughly 60\% lower than each statistical benchmark. DM statistics against the post-intervention configuration are uniformly large and negative ($-23.74$ random walk, $-22.41$ AR(1), $-21.86$ ARMA-GARCH; all $p < 10^{-6}$). ARMA-GARCH is only marginally more accurate than AR(1) on the mean forecast, since conditional-variance machinery does not change the mean equation. The pre-intervention agentic system already outperforms each benchmark by a wide margin (consistent with \cite{alridhawi2026regime}); the closed-loop yields a further, statistically significant improvement on top of an already-strong baseline.

\paragraph{Conditional predictive ability.} The pooled DM test evaluates unconditional equal predictive accuracy and can mask asymmetric performance. We therefore apply the conditional predictive ability test of \cite{giacomini2006tests} with the autoencoder regime label $\ell_t$ as conditioning variable. Regressing $d_{i,t} = \beta_0 + \beta_1 \ell_t + \varepsilon_{i,t}$ and testing $\beta_0 = \beta_1 = 0$ via a Wald statistic with HAC-corrected standard errors, two-way fixed-effects pooling yields Wald $= 86.41$ ($p < 10^{-9}$), with $\hat\beta_0 = 0.014$ ($t = 4.27$) and $\hat\beta_1 = 0.029$ ($t = 5.93$). Post-intervention loss is lower in both regimes (differential 0.014 low-vol., 0.043 high-vol.); the advantage is roughly three times larger in high volatility, matching the behavioral analysis: gains are largest where pre-intervention deficiencies in risk calibration and adaptation were most severe.

\paragraph{Model confidence set across closed-loop cycles.} The Hansen Model Confidence Set \cite{hansen2011model} is computed on test-period squared-error losses for five candidates: pre-intervention, cycle~1, cycle~2, cycle~3, and a non-corrective ablation matching cycle~3 in additional SAC training steps but with $\lambda = 0$ (no LLM penalty). The $T_{\max}$ statistic with 5{,}000 stationary block-bootstrap resamples (block length 10 days) gives: 90\% MCS contains $\{$cycle~3, cycle~2$\}$; 75\% MCS shrinks to $\{$cycle~3$\}$. The $\lambda = 0$ ablation is rejected at 99\% with $T_{\max} = 11.62$ ($p < 0.001$) against cycle~3, confirming that the predictive-accuracy gains require the LLM penalty rather than generic additional training. Pre-intervention and cycle~1 are rejected from the 90\% MCS with $T_{\max}$ of $9.07$ and $3.62$, tracking the per-cycle behavioral progression.

\paragraph{Robustness across loss functions.} Forecast rankings can reverse under different losses, so the DM test is repeated under absolute-error loss $|e_{i,t}|$, MAPE loss $|e_{i,t}|/|y_{i,t}|$, and the QLIKE loss $\log(\sigma^{2}_{i,t}) + e^{2}_{i,t}/\sigma^{2}_{i,t}$ with $\sigma^{2}_{i,t}$ a 22-day realized-variance proxy. Pooled DM statistics are $-7.83$ (squared-error), $-7.21$ (MAE), $-7.58$ (MAPE), $-7.04$ (QLIKE), all $p < 0.001$. The narrow spread indicates the improvement is not an artifact of the squared-error metric.

\subsection{Framework Ablations}\label{sec:ablation}

Three ablations isolate the contribution of individual design choices: the number of LLM judges, the credit assignment strategy, and the sensitivity of scores to prompt phrasing. Each modifies a single framework component while keeping the others fixed, and the closed-loop cycle is re-run from the same pre-intervention checkpoint.

\subsubsection{Number of judges}

To assess whether the three-judge ensemble is necessary, the closed-loop process was repeated with each judge in isolation, holding all other components fixed. Per-episode composite-score variance is the diagnostic of interest because a noisier consensus mean produces noisier SAC updates. Table~\ref{tab:judge_ablation} reports the resulting prediction quality and score-variance figures.

\begin{table}[!t]
\centering
\caption{Closed-loop outcomes after three evaluation cycles for single-judge variants versus the three-judge ensemble. Score variance is the mean per-episode composite-score standard deviation across the 200 evaluation episodes.}
\label{tab:judge_ablation}
\footnotesize
\setlength{\tabcolsep}{3pt}
\begin{tabular}{@{}lcccc@{}}
\toprule
Judge Configuration & MAPE (\%) & DA (\%) & Sharpe & Score Variance \\
\midrule
GPT~5.4 only          & 0.55 & 74 & 1.57 & 0.41 \\
Claude~4.6 Opus only  & 0.56 & 74 & 1.53 & 0.43 \\
Gemini~3.1 Pro only   & 0.58 & 73 & 1.46 & 0.46 \\
3-judge ensemble      & 0.54 & 74 & 1.61 & 0.29 \\
\bottomrule
\end{tabular}
\end{table}

All single-judge variants improve over the pre-intervention baseline, so the closed-loop mechanism is not dependent on any single LLM family. The three-judge ensemble nevertheless achieves consistently better outcomes; its primary advantage is variance reduction (per-episode variance 0.29 is 29 to 37\% lower than any single judge), which translates into a more stable reward signal for SAC training. GPT~5.4 comes closest as a single judge but its higher score variance produces noisier updates and a lower final Sharpe. The single-judge MAPE gap to the ensemble is 0.01 to 0.04, indicating robustness to the choice of judge with a measurable ensemble benefit through variance reduction.

\subsubsection{Credit assignment strategy}

An alternative uniform strategy was tested in which all dimension penalties are applied equally to both action components, removing the dimension-to-action mapping in Table~\ref{tab:credit}. Table~\ref{tab:credit_ablation} compares the two strategies.

\begin{table}[!t]
\centering
\caption{Comparison of targeted versus uniform credit assignment after three evaluation cycles. $\Delta \bar{s}_{\text{def}}$ and $\Delta \bar{s}_{\text{non}}$ denote mean score changes for initially deficient and non-deficient dimensions, respectively.}
\label{tab:credit_ablation}
\footnotesize
\setlength{\tabcolsep}{3pt}
\begin{tabular}{@{}lcccc@{}}
\toprule
Credit Strategy & MAPE (\%) & DA (\%) & $\Delta \bar{s}_{\text{def}}$ & $\Delta \bar{s}_{\text{non}}$ \\
\midrule
Targeted (proposed) & 0.54 & 74 & +1.3 & +0.1 \\
Uniform             & 0.56 & 73 & +0.9 & $-$0.4 \\
\bottomrule
\end{tabular}
\end{table}

The targeted strategy produces larger improvements on deficient dimensions ($+$1.3 versus $+$0.9 points) while preserving non-deficient ones (mean change $+$0.1). The uniform strategy degrades non-deficient dimensions by 0.4 points on average because uniform penalties create conflicting gradients: a penalty intended to improve routing inappropriately adjusts the anomaly threshold, degrading previously adequate regime detection. In this setting, targeted credit assignment is therefore necessary for translating LLM behavioral assessments into non-destructive learning signals.

\subsubsection{Prompt sensitivity}

To probe sensitivity of scores to prompt phrasing, three prompt variants were constructed: the original prompt (A), a version with reordered dimension descriptions (B), and a version with simplified language preserving the rubric criteria (C). Each was used with GPT~5.4 to evaluate the same 60 unperturbed validation-period episodes, with per-dimension scores compared across variants. The assessment is confined to the validation period because any sensitivity finding that informed a prompt revision would otherwise constitute indirect selection on the test data.

This is a bounded robustness check rather than a settled claim of prompt invariance: it covers 60 episodes, a single judge (GPT~5.4), and only two surface-level variant axes (ordering and lexical simplification). A definitive claim would require a larger pool, the full ensemble, and more variant types (paraphrased rubrics, alternative chain-of-thought scaffolds, adversarial phrasings).

Within these limits, the mean absolute score difference between any two variants is 0.14 per dimension on the 1 to 5 scale, with an intra-class correlation coefficient (ICC) of 0.91. No dimension shows systematic bias; the largest per-dimension shift is 0.11 (strategy coherence, Variant C), within the standard error of 0.09. The ICC exceeds the 0.75 threshold conventionally considered ``excellent'' \cite{cicchetti1994guidelines}, indicating scores are driven primarily by trace properties rather than phrasing differences. The reordering test (Variant B) specifically probes position bias; the absence of an effect ($p = 0.34$, paired $t$-test on composite scores) is consistent with robustness to this class of artifact within the tested range.

%========================================================================
\section{Discussion}\label{sec:discussion}
%========================================================================

\subsection{Interpretation of Findings}\label{sec:interpretation}

The empirical results support a three-part claim about LLM-based behavioral evaluation of agentic AI systems. The dimensions can be assessed with selective sensitivity rather than as a diffuse quality signal: each of the six perturbations produced a targeted score drop of at least 1.6 points while shifting the remaining dimensions by an average of only 0.32 points. This asymmetry requires that the judges correctly identify which trace component is responsible for the degradation and distinguish a targeted deficiency from its downstream consequences. The cross-model agreement (Krippendorff's $\alpha$ 0.74 to 0.85) indicates the selective sensitivity is not an artifact of any single model.

The scores also carry information beyond what is visible at the trace level: the composite correlates at $\rho = 0.72$ with realized 20-day Sharpe ratio, exceeding any individual dimension's correlation. The distinct predictive profiles (regime detection predicting MAPE, risk calibration predicting drawdown, strategy coherence predicting returns) align with how each behavioral property should affect outcomes, supplying convergent validity. The qualitative analysis (Section~\ref{sec:qualitative}) shows the mechanism: the judges ground scores in specific numerical trace values and articulate causal chains between upstream decisions and downstream consequences.

The assessments are not merely diagnostic but actionable. The 11.5\% one-day MAPE reduction (significant under paired $t$-test, Diebold-Mariano, and across four loss specifications) and the 3-point directional-accuracy gain are meaningful improvements over an already-strong baseline. Several observations support a causal interpretation: the gains concentrate in the dimensions identified as initially deficient; they are largest in high-volatility episodes where deficiencies were most severe, consistent with targeted correction rather than uniform improvement; non-targeted dimensions remained stable, ruling out generic SAC re-training as the source of the gains; and the Hansen MCS rejects a non-corrective $\lambda = 0$ ablation (same SAC training steps without LLM penalty) at the 99\% level against the closed-loop configuration.

The ablations sharpen these conclusions. The multi-judge ensemble reduces per-episode score variance by 29 to 37\%, producing a more stable reward signal, though the modest single-judge gap (0.01 to 0.04 MAPE) indicates the framework is robust to the choice of judge. The credit-assignment ablation is more consequential: uniform penalty distribution degrades non-targeted dimensions by 0.4 points on average, so targeted credit assignment is necessary for translating behavioral diagnostics into non-destructive learning signals. The $\lambda$ sweep (Table~\ref{tab:lambda}) maps a clean transition from under-correction ($\lambda < 0.10$) through an effective range ($\lambda \in [0.10, 0.20]$) to destabilization ($\lambda > 0.25$), supplying concrete tuning guidance.

\subsection{Implications for Applied AI and Agentic-System Evaluation}\label{sec:implications}

The findings carry three implications that extend beyond the financial-forecasting demonstration, each addressing a different layer of the methodology stack: how agentic process quality is measured, how the LLM-as-a-Judge paradigm extends to temporal decision sequences, and how qualitative diagnostics translate into a quantitative reinforcement-learning signal. All three transfer to any application that records intermediate decisions in a structured form.

For agentic-system evaluation generally, the results show that process quality can be assessed directly (rather than inferred indirectly from output success) with sufficient reliability to support both diagnostic reporting and downstream optimization, provided three design conditions are met: intermediate decisions must be loggable in a structured format; evaluation dimensions must be domain-specific rather than borrowed wholesale from natural-language evaluation; and dimension specificity must be validated through engineered perturbations rather than assumed. The methodology developed here satisfies all three and produces a quantitative answer to a question aggregate metrics cannot address: which intermediate decision was unsound when the system failed.

For the LLM-as-a-Judge paradigm, the framework extends scoring from static outputs \cite{zheng2023judging,liu2023geval,dubois2024alpacafarm} to temporal sequences of interdependent decisions under stochastic conditions, where decision quality cannot always be inferred from immediate outcomes. The structured prompt and chain-of-thought scaffold from \cite{liu2023geval} transfer without loss of judge agreement; what is added is the episode formalization, the trace serialization, and the credit-assignment routing. The extension is not specific to finance: autonomous driving (with intermediate perception and planning decisions), medical decision support (with intermediate diagnostic and treatment-selection decisions), and robotic control (with intermediate state estimation and action selection) are all candidate domains.

For the bridge from qualitative LLM evaluation to quantitative reinforcement learning, the closed-loop mechanism extends the RLHF principle \cite{ouyang2022training,dubois2024alpacafarm} (originally developed for language-model alignment) to a setting in which the learning signal must be routed to specific components of a controller's action space. The credit-assignment mechanism is the technical contribution that makes this routing possible. The targeted-versus-uniform ablation quantifies the value: uniform penalties degrade non-targeted behaviors by 0.4 points; targeted penalties improve targeted behaviors by 1.3 points while keeping non-targeted ones stable. Practitioners implementing analogous closed-loop systems will likely face the same routing problem.

\subsection{Generalizability beyond the Demonstration Domain}\label{sec:generalizability}

The methodology, considered separately from the empirical case study, is application-agnostic. The four core components transfer without modification: the behavioral-trace formalization (Equation~\ref{eq:trace}), the episodic grouping (Equation~\ref{eq:episode}), the perturbation-based validation (Section~\ref{sec:perturb_design}), and the credit-assignment mechanism (Section~\ref{sec:credit}). What requires redesign for a new domain is the specific list of dimensions, their rubric anchors, and the action-subspace mapping, all of which depend on the architecture of the system being evaluated.

The empirical findings are quantitative artefacts of the specific case study and should be re-estimated elsewhere. The 11.5\% MAPE reduction, the $\rho = 0.72$ predictive-validity correlation, the regime localization, and the specific Krippendorff's $\alpha$ values reflect properties of the AE-NodeFormer + SAC architecture, the 20 S\&P~500 equities, and the 2017 to 2025 test window. The directionality (perturbations produce targeted drops, scores correlate with realized performance, closed-loop intervention improves prediction quality) should transfer to analogous agentic systems, but the magnitudes should not be assumed without re-evaluation. Priority extensions include macroeconomic nowcasting, energy demand forecasting under regime shifts, supply-chain forecasting under demand-disruption events, autonomous decision systems in industrial control, and medical decision-support pipelines.

\subsection{Limitations}\label{sec:limitations}

LLM judges are generalist models whose domain knowledge derives from pre-training rather than domain-specific training. The structured prompt and rubric mitigate this by defining criteria explicitly, but the judges may miss subtle signals a human expert would catch (options-implied volatility surfaces, cross-asset contagion patterns) that are not represented in the trace fields. The predictive-validity correlations suggest assessments are meaningful but possibly incomplete relative to expert human evaluators; a direct LLM-vs-human comparison would strengthen the validation and is beyond the present scope.

The closed-loop mechanism introduces a Goodhart-type objective-misspecification risk: the system could optimize for LLM approval rather than genuine task performance. No such effect was observed (all dimension improvements translated into prediction gains, non-targeted dimensions remained stable), and the $\lambda$ sensitivity analysis (Table~\ref{tab:lambda}) bounds the operating range within which LLM feedback improves performance. The periodic evaluation protocol, which restricts feedback to corrective interventions rather than continuous reward modification, supplies a structural safeguard. The risk nevertheless remains a theoretical concern under many more cycles or on systems whose behavioral patterns diverge substantially from pre-training.

The framework has been validated on a single architecture (the AE-NodeFormer + SAC system from \cite{alridhawi2026regime}) and a single case study (20 S\&P~500 equities, 2{,}267 trading days from 2017 to 2025). The 20-equity sample and 8-year horizon were chosen to keep LLM evaluation cost tractable. The test window contains the 2018 volatility shock, the 2020 COVID-19 dislocation, and the 2022 to 2023 inflation regime, providing exposure to structural breaks, but the empirical magnitudes should still be re-estimated on architecturally distinct systems and broader samples. The LLM evaluation cost (approximately \$180) is modest in absolute terms but scales linearly with episodes and judges; real-time deployment or higher-frequency cycles may require distillation of the judges into a smaller, faster surrogate, subject to validation that distillation preserves dimension specificity and predictive validity.

The three judges are commercial systems whose weights are subject to provider-side updates the authors cannot version-pin from outside the API. Mitigations include temperature-zero querying, the verbatim frozen prompt in Appendix~\ref{app:prompts}, the three-family ensemble (reducing any single provider's update influence), and an archive of per-episode dimension scores and rationales for the entire 2017 to 2025 test window. A contemporaneous replay against the archive is therefore reproducible; a future re-query against the live APIs may diverge if any provider has updated its model. The inter-judge agreement (Krippendorff's $\alpha$ 0.74 to 0.85) suggests the framework is not critically dependent on any single provider, so partial drift would be partially absorbed; complete drift across all three would require re-validation.

Finally, temperature-zero operation produces deterministic evaluations but eliminates the stochastic variation that could reveal model uncertainty on borderline cases. Extending the framework to incorporate evaluation uncertainty (for example via multi-temperature sampling and per-episode score variance) could soften credit-assignment penalties on borderline episodes and is left to future work.

\subsection{Future Work}\label{sec:future}

Three directions extend the present study. Validation on architecturally distinct agentic systems in non-financial domains (industrial control, medical decision support, energy demand forecasting under regime shifts) would test the application-agnostic claim directly; the trace formalization, perturbation procedure, predictive-validity check, and credit-assignment routing all transfer in principle, but specific dimensions and rubric anchors will differ.

Distillation of the multi-judge ensemble into a faster surrogate is the second direction. A distilled model that preserves dimension specificity and predictive validity would reduce evaluation latency and cost, opening the framework to real-time deployment. A multi-headed distillation, in which the student outputs one score per dimension trained against all three judges' scores jointly, is a natural starting point.

Incorporation of evaluation uncertainty through temperature sampling is the third direction. Per-episode uncertainty estimates could soften credit-assignment penalties on borderline episodes (mitigating the Goodhart-type risk when the LLM is treated as a deterministic oracle) and flag episodes for human review when the judges agree on a low score but disagree on the failure label, precisely the case where intervention is most consequential.

%========================================================================
\section*{Disclosure of interest}
%========================================================================

The authors report there are no competing interests to declare. No financial or personal relationships with the providers of the language models used as evaluators, or with any other party that could be perceived to have an interest in the empirical findings, exist for any of the authors.

%========================================================================
\section*{Funding}
%========================================================================

The authors received no financial support for the research, authorship, or publication of this article.

%========================================================================
\section*{Data availability statement}
%========================================================================

Daily price and volume data for the twenty S\&P~500 equities analyzed in this study, together with the CBOE Volatility Index (VIX) used as a market-volatility proxy, were retrieved from Yahoo Finance through the \texttt{yfinance} Python library. The retrieved snapshot has been deposited at Zenodo under DOI 10.5281/zenodo.20192822 (\url{https://doi.org/10.5281/zenodo.20192822}, Creative Commons Attribution 4.0 International License) and includes the per-ticker daily OHLCV records together with the script used to generate them. The sentiment series was constructed from public posts on X (formerly Twitter) collected through cashtag-based queries against the X application programming interface. The X developer agreement restricts redistribution of platform content and of content derived from it; the raw posts and the aggregated daily sentiment series are therefore not redistributable, and reproduction requires independent collection under the platform terms in force at the time of access. The code implementing the behavioral evaluation framework, the perturbation harness, and the closed-loop reward integration is available from the corresponding author on reasonable academic request.

%========================================================================
\bibliographystyle{IEEEtran}
\bibliography{references}
%========================================================================

%========================================================================
\appendix
\renewcommand{\thetable}{A\arabic{table}}
\renewcommand{\thefigure}{A\arabic{figure}}
\setcounter{table}{0}
\setcounter{figure}{0}
\section{LLM Judge Prompt Template}\label{app:prompts}
%========================================================================

This appendix records the prompt template applied uniformly to all three judges (GPT~5.4, Claude~4.6 Opus, and Gemini~3.1 Pro) at temperature zero with a $2{,}000$-token output budget. A fixed system message establishes the evaluator role, trace semantics, and rubric; an episode-specific behavioral-trace serialization is appended; the model is constrained by an output schema together with the failure-label vocabulary in Table~\ref{tab:failures}. Only the trace serialization varies across episodes.

\subsection*{A.1 System message}\label{app:system}

The text below is delivered as the system role (or its provider-specific equivalent) on every episode evaluation.

{\tiny
\begin{verbatim}
You are an expert evaluator of an agentic stock prediction system that combines an
autoencoder regime detector, a routing layer that selects between a normal pathway
and an event pathway, dual node-transformer prediction pathways, and a Soft Actor-
Critic (SAC) controller that adjusts the regime threshold tau and the blending
weight alpha at each trading day.

For each episode you receive five consecutive daily behavioral traces. Each trace
contains: (i) market context (closing price, volume, VIX, aggregated BERT
sentiment); (ii) autoencoder reconstruction error e, current threshold tau, and
regime label (0 = normal, 1 = anomalous); (iii) routing blending weight alpha and
dominant pathway index; (iv) SAC adjustments delta_tau and delta_alpha, each in
[-0.1, 0.1]; (v) the one-day-ahead prediction; (vi) trailing 20-day MAPE and
directional accuracy.

Interpretation rules. A regime label of 1 is appropriate when the reconstruction
error exceeds the threshold by a non-trivial margin and the surrounding market
context (VIX above 25, sharply negative sentiment, abrupt price moves) supports
the anomaly classification. A blending weight alpha closer to 1 directs prediction
weight to the normal pathway; alpha closer to 0 directs it to the event pathway.
SAC adjustments are applied additively at the next time step. Adaptive behavior
should be timely (within one to two days of a detectable change), proportional to
the magnitude of that change, and stable (no day-to-day reversals).

Score each episode along the six dimensions (RD, RT, AD, RC, SC, ER) on an integer
1-5 scale: 1 = fundamentally flawed; 2 = predominantly flawed with occasional
acceptable behavior; 3 = acceptable with identifiable weaknesses; 4 = strong with
minor imperfections; 5 = exemplary.

The score-1 and score-5 anchors for each dimension are:
  RD  1: systematic misclassification of regime; threshold unresponsive to
         volatility.
      5: all regime transitions identified within one trading day; threshold
         adjustments proportional to volatility.
  RT  1: data routed to wrong pathway; blending weight contradicts regime
         classification.
      5: routing consistently matches market state; smooth blending transitions
         near threshold.
  AD  1: parameters frozen or wildly oscillating; no response to condition
         changes within episode.
      5: prompt, proportional adjustments within one to two days; no overshoot
         or oscillation.
  RC  1: risk posture inappropriate for volatility (e.g., aggressive in high-VIX
         period).
      5: conservative during high volatility, confident during low volatility;
         smooth transitions.
  SC  1: multiple contradictory actions across the decision chain within episode.
      5: all decisions form a logically consistent sequence; no contradictions.
  ER  1: no corrective action within two days of an error; errors persist or
         worsen.
      5: corrective adjustments within one to two days; subsequent predictions
         show measurable improvement.
Intermediate levels interpolate between the anchors using the per-dimension rubric
descriptions provided to the judge.

Reason in four steps before scoring: (1) summarize market conditions across the
five-day window, noting any volatility, sentiment, or price transitions; (2)
examine each decision point and assess whether the decision was appropriate given
the information available at that time; (3) identify any inconsistencies,
failures, or suboptimal behaviors, citing the specific trace fields that support
each observation; (4) assign each of the six scores, grounded in specific
observations from Steps 2 and 3, not an overall impression.

Return a single JSON object in the format specified in the output schema. Do not
include the chain-of-thought reasoning inside the JSON.
\end{verbatim}
}

\subsection*{A.2 Behavioral-trace serialization}\label{app:trace}

Each daily trace $\mathcal{B}_t$ defined in Equation~\ref{eq:trace} is serialized as a JSON record and the five traces of an episode are presented in chronological order; a representative high-volatility record is shown below.

{\tiny
\begin{verbatim}
{
  "day_index": 3,
  "market_context": {"price": 142.31, "volume": 38421500, "VIX": 22.4,
                     "sentiment_bar": -0.18},
  "autoencoder": {"reconstruction_error_e": 0.034, "threshold_tau": 0.031,
                  "ratio_e_over_tau": 1.10, "regime_label": 1},
  "routing": {"alpha": 0.35, "normal_pathway_pct": 35,
              "event_pathway_pct": 65, "dominant_pathway": "event"},
  "sac_action": {"delta_tau": -0.002, "delta_alpha": -0.05,
                 "tau_after_update": 0.029, "alpha_after_update": 0.30},
  "prediction": {"y_hat_next_day": 142.05},
  "rolling_performance": {"MAPE_20d_pct": 0.58, "DA_20d_pct": 73,
                          "trend": "stable"}
}
\end{verbatim}
}

\subsection*{A.3 Output schema and failure-label vocabulary}\label{app:output}

The judge returns a single JSON object with one integer score per dimension (1 to 5), one justification per dimension referencing specific trace fields, and an optional failure label for each dimension scoring strictly below $\theta = 3$.

{\tiny
\begin{verbatim}
{
  "scores": {"RD": <int>, "RT": <int>, "AD": <int>,
             "RC": <int>, "SC": <int>, "ER": <int>},
  "justifications": {"RD": "<text referencing trace fields>",
                     "RT": "<text>", "AD": "<text>",
                     "RC": "<text>", "SC": "<text>", "ER": "<text>"},
  "failures": [{"dimension": "<RD|RT|AD|RC|SC|ER>",
                "label": "<label-from-vocabulary>"}]
}
\end{verbatim}
}

The 12 failure labels and their default SAC action subspaces are listed in Table~\ref{tab:failures}; a label is recorded only when the corresponding score is strictly below $\theta = 3$, and determines which action component receives the penalty under the credit-assignment routing of Table~\ref{tab:credit}.

\begin{table}[H]
\centering
\caption{Failure-label vocabulary and default SAC action-subspace mapping. The default mapping follows Table~\ref{tab:credit}; a label may override it in atypical failure modes (Section~\ref{sec:credit}).}
\label{tab:failures}
\footnotesize
\setlength{\tabcolsep}{5pt}
\begin{tabular}{lcc}
\toprule
\textbf{Label} & \textbf{Dimension} & \textbf{Subspace} \\
\midrule
\texttt{delayed\_threshold}            & RD & $\Delta\tau$ \\
\texttt{systematic\_misclassification} & RD & $\Delta\tau$ \\
\texttt{oversensitive\_threshold}      & RD & $\Delta\tau$ \\
\texttt{wrong\_routing}                & RT & $\Delta\alpha$ \\
\texttt{inconsistent\_blend}           & RT & $\Delta\alpha$ \\
\texttt{abrupt\_blending}              & RT & $\Delta\alpha$ \\
\texttt{frozen\_parameters}            & AD & $\Delta\tau,\,\Delta\alpha$ \\
\texttt{oscillating\_actions}          & AD & $\Delta\tau,\,\Delta\alpha$ \\
\texttt{uncalibrated\_risk}            & RC & $\Delta\tau$ \\
\texttt{contradictory\_decisions}      & SC & $\Delta\tau,\,\Delta\alpha$ \\
\texttt{delayed\_recovery}             & ER & $\Delta\tau,\,\Delta\alpha$ \\
\texttt{error\_amplification}          & ER & $\Delta\tau,\,\Delta\alpha$ \\
\bottomrule
\end{tabular}
\end{table}

%========================================================================
\begin{IEEEbiography}[{\includegraphics[width=1in,height=1.25in,clip,keepaspectratio]{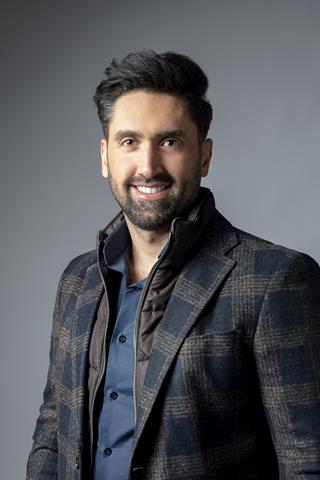}}]{Mohammad Al Ridhawi}
received the B.A.Sc.\ degree in computer engineering and the M.Sc.\ degree in digital transformation and innovation (machine learning) from the University of Ottawa, Ottawa, Canada, in 2019 and 2021, respectively. He is currently pursuing the Ph.D.\ degree in electrical and computer engineering at the University of Ottawa, where he also serves as a Part-Time Engineering Professor. He has industry experience as a Senior Data Scientist and Senior Machine Learning Engineer, building production ML systems in financial and environmental domains. His research interests include deep learning, graph neural networks, natural language processing, financial time series analysis, and reinforcement learning.
\end{IEEEbiography}

\begin{IEEEbiography}[{\includegraphics[width=1in,height=1.25in,clip,keepaspectratio]{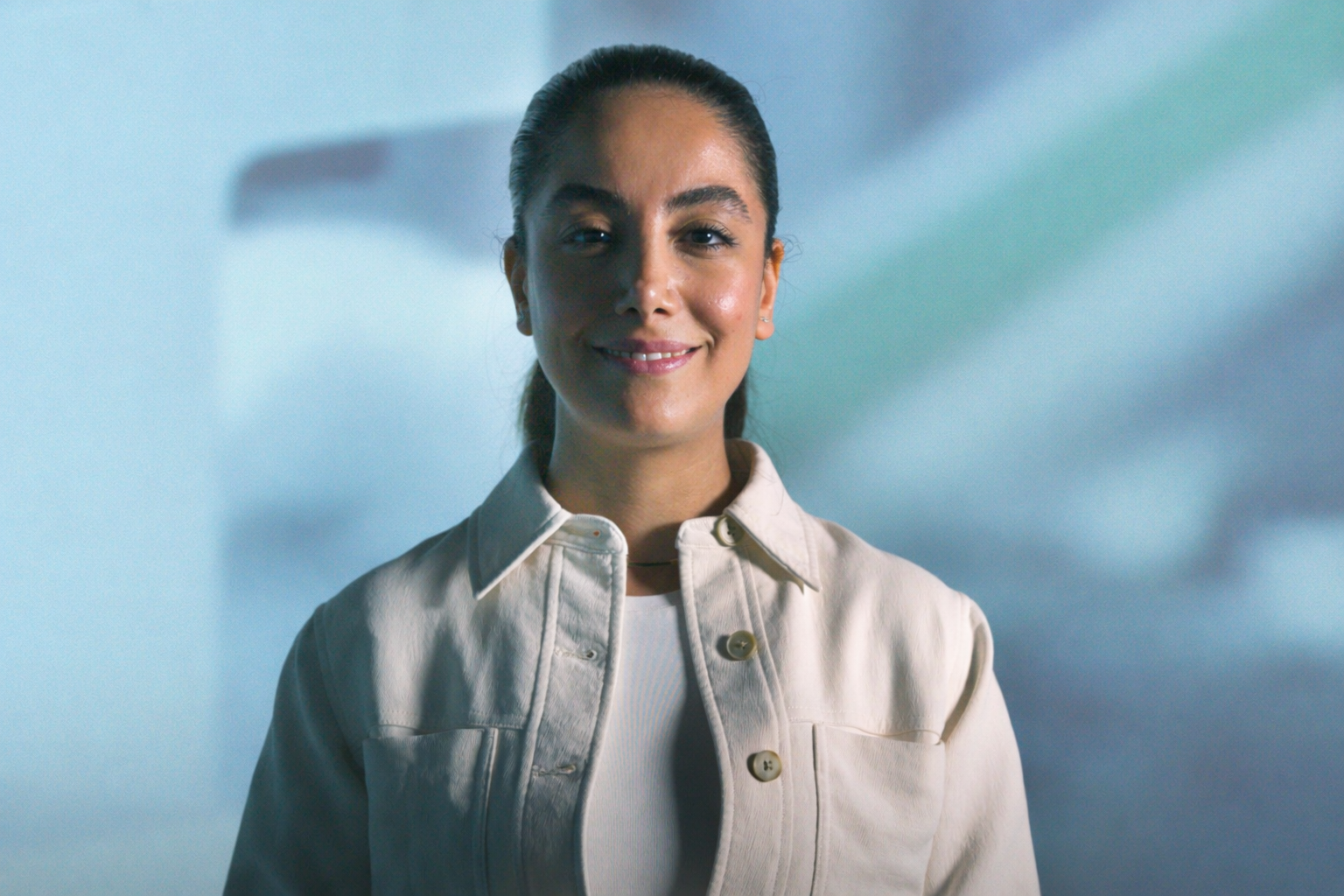}}]{Mahtab Haj Ali}
received the M.Sc.\ degree in digital transformation and innovation from the University of Ottawa, Ottawa, Canada, in 2021. She is currently pursuing the Ph.D.\ degree in electrical and computer engineering at the University of Ottawa, with a research focus on time series forecasting and deep learning models. She works as an AI Research Engineer at the National Research Council of Canada, where she builds and evaluates large language models (LLMs) and develops AI-driven solutions for real-world industrial applications. Her work includes large-scale time series analysis, advanced feature engineering, and the application of LLMs in production environments. Her research interests include deep learning for time series analysis, deep neural networks, and applied artificial intelligence.
\end{IEEEbiography}

\begin{IEEEbiography}[{\includegraphics[width=1in,height=1.25in,clip,keepaspectratio]{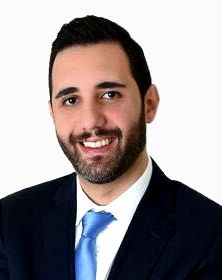}}]{Hussein Al Osman}
received the B.A.Sc., M.A.Sc., and Ph.D.\ degrees from the University of Ottawa, Ottawa, Canada. He is an Associate Professor and Associate Director in the School of Electrical Engineering and Computer Science at the University of Ottawa, where he leads the Multimedia Processing and Interaction Group. His research focuses on affective computing, multimodal affect estimation, human--computer interaction, serious gaming, and multimedia systems. He has produced over 50 peer-reviewed research articles, two patents, and several technology transfers to industry.
\end{IEEEbiography}

\EOD

\end{document}